\documentclass[10pt,twocolumn,letterpaper]{article}

\usepackage{iccv}
\usepackage{times}
\usepackage{epsfig}
\usepackage{graphicx}
\usepackage{amsmath}
\usepackage{amssymb}

\usepackage{amsfonts, mathtools, algorithm, algorithmic}
\usepackage{color, xcolor, colortbl, graphicx}
\usepackage{multirow, subfigure, wrapfig, caption, booktabs}
\usepackage{listings, enumitem, caption}

\newcommand{\method}{{Occlusion Completion}}
\newcommand{\abbrev}{{OcCo\,}}
\newcommand{\Pc}{{\mathcal{P}}}

\makeatletter
\newcommand{\subalign}[1]{%
  \vcenter{%
    \Let@ \restore@math@cr \default@tag
    \baselineskip\fontdimen10 \scriptfont\tw@
    \advance\baselineskip\fontdimen12 \scriptfont\tw@
    \lineskip\thr@@\fontdimen8 \scriptfont\thr@@
    \lineskiplimit\lineskip
    \ialign{\hfil$\m@th\scriptstyle##$&$\m@th\scriptstyle{}##$\hfil\crcr
      #1\crcr
    }%
  }%
}
\makeatother

\definecolor{dkgreen}{rgb}{0,0.6,0}
\definecolor{gray}{rgb}{0.5,0.5,0.5}
\definecolor{mauve}{rgb}{0.58,0,0.82}

\definecolor{CRed}{HTML}{EF476F}
\definecolor{CYellow}{HTML}{FFD166}
\definecolor{CBlue}{HTML}{118AB2}

\newsavebox{\codebox}

\definecolor{citecolor}{HTML}{0071bc}
\usepackage[pagebackref=true,breaklinks=true,colorlinks,citecolor=citecolor,bookmarks=false]{hyperref}

\iccvfinalcopy 

\def\iccvPaperID{1727} 

\ificcvfinal\fi

\begin{document}

\title{Unsupervised Point Cloud Pre-training via Occlusion Completion}

\author{
  Hanchen Wang$^{1}$ \quad\quad Qi Liu$^{2}$ \quad\quad Xiangyu Yue$^{3}$ \quad\quad Joan Lasenby$^{1}$ \quad\quad Matt J. Kusner$^{4}$\\\vspace{2ex}
  {\normalsize $^1$University of Cambridge \quad\quad $^2$University of Oxford \quad\quad $^3$UC Berkeley \quad\quad $^4$University College London}
}

\maketitle
\ificcvfinal\thispagestyle{empty}\fi

\begin{abstract}\vspace{-.3cm}
   We describe a simple pre-training approach for point clouds. It works in three steps: 1. Mask all points occluded in a camera view; 2. Learn an encoder-decoder model to reconstruct the occluded points; 3. Use the encoder weights as initialisation for downstream point cloud tasks. We find that even when we pre-train on a single dataset (ModelNet40), this method improves accuracy across different datasets and encoders, on a wide range of downstream tasks. 
   Specifically, we show that our method outperforms previous pre-training methods in object classification, and both part-based and semantic segmentation tasks. 
   We study the pre-trained features and find that they lead to wide downstream minima, have high transformation invariance, and have activations that are highly correlated with part labels. 
   Code and data are available at: \url{https://github.com/hansen7/OcCo}
\end{abstract}
\vspace{-2ex}
\section{Introduction}
\label{sec:intro}
There has been a flurry of exciting new point cloud models for object detection~\cite{lang2019pointpillars,wang2020pillar,zhou2018voxelnet} and segmentation~\cite{hu2019randla,landrieu2018large,yang2019learning,zhu2020cylindrical}. These methods rely on large scale point cloud datasets that are labelled. Unfortunately, labelling point clouds is challenging for a number of reasons: 
(1) Point clouds can be sparse, occluded, and at low resolutions, making the identity of points ambiguous; 
(2) Datasets that are not sparse can easily reach hundreds of millions of points (e.g., small dense point clouds for object classification~\cite{zhou2013complete} and large vast point clouds for reconstruction~\cite{zolanvari2019dublincity});
(3) Labelling individual points or drawing 3D bounding boxes are both more time-consuming and error-prone than labelling 2D images~\cite{wang2019latte}. 
These challenges have impeded the deployment of point cloud models into new real world settings where labelled data is scarce. 


\begin{figure}[t!]
    \centering
    \includegraphics[width=\columnwidth]{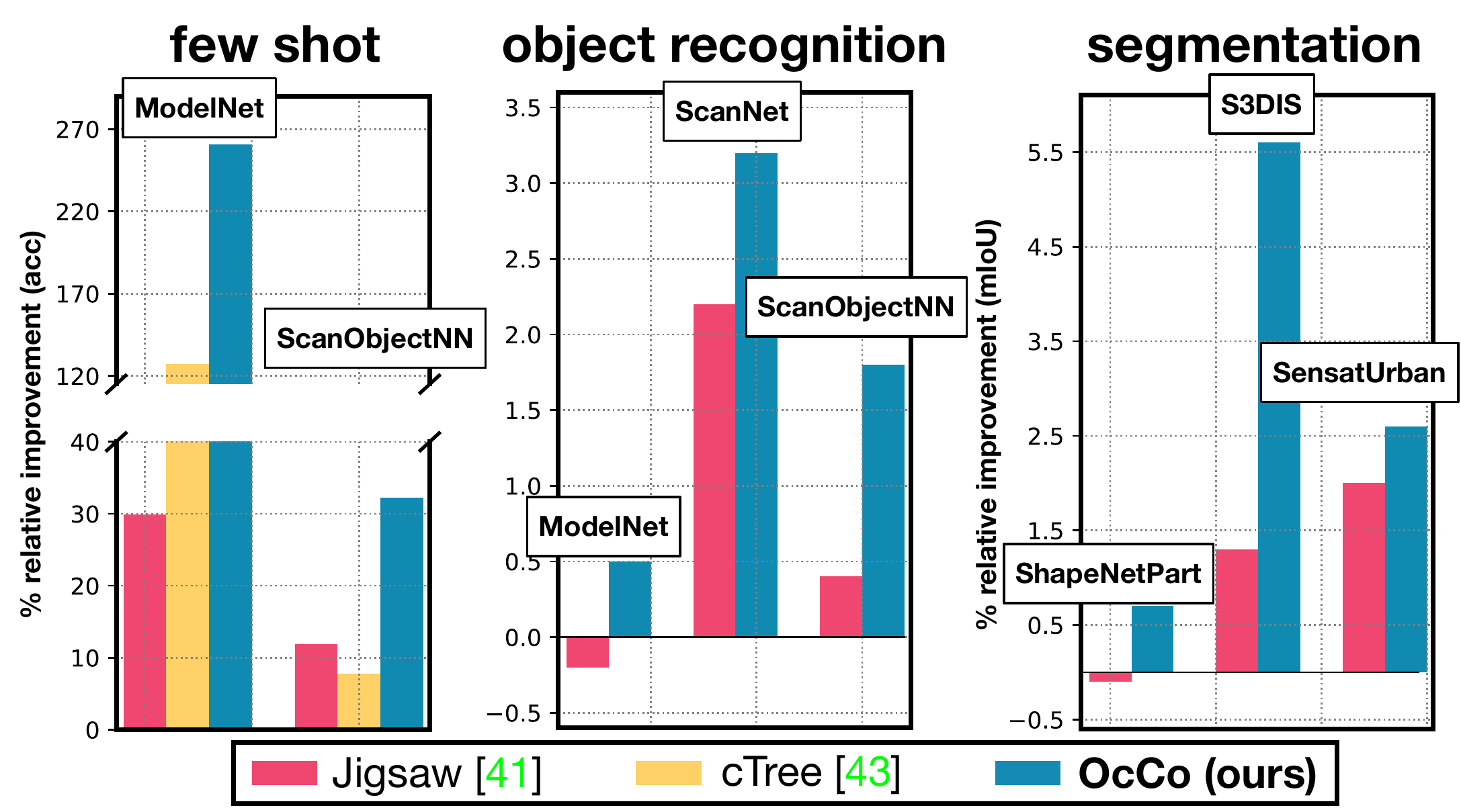}
    \vspace{-3ex}
    \caption{The relative improvement over random initialisation of multiple pre-training methods: {Jigsaw}~\cite{sauder2019self}, {cTree}~\cite{sharma2020self}, and {OcCo} (ours) for various downstream tasks.}
    \label{fig:rel_imp}
    \vspace{-2ex}
\end{figure}

However, current 3D sensing modalities (\textit{i.e.}, 3D scanners, stereo cameras, lidars) have enabled the creation of large \emph{unlabelled} repositories of point cloud  data~\cite{hackel2017isprs,Rusu_ICRA2011_PCL}. 
This has inspired a recent line of work on 
unsupervised pre-training methods to learn point cloud model initialisation. Initial work used latent generative models such as generative adversarial networks (GANs)~\cite{achlioptas2017learning,han2019view,wu2016learning} and autoencoders~\cite{hassani2019unsupervised,li2018so,yang2018foldingnet}. These have been recently outperformed by self-supervised  objectives~\cite{sauder2019self,pointcontrast,alliegro2020joint,sharma2020self,hou2020exploring,zhang2021self}.
 

Inspired by this recent line of work, we propose \emph{\method} (OcCo), an unsupervised pre-training method that consists of: (a) a mechanism to generate masked point clouds via view-point occlusions, and (b) a completion task to reconstruct the occluded point cloud. The idea of occlusion+completion is grounded in three observations: (1) A pre-trained model that is accurate at completing occluded point clouds needs to understand spatial and semantic properties of these point clouds. (2) 3D scene completion~\cite{song2017semantic,dai2020sg,hou2020revealnet} has been shown to be a useful auxiliary task for learning representations for visual localisation~\cite{schonberger2018semantic}. (3) Mask-based completion tasks have become the \textit{de facto} standard for learning pre-trained representations in natural language processing~\cite{devlin2018bert,mikolov2013efficient,peters2018deep} and are widely used in pre-training for images~\cite{inpainting} and graphs~\cite{GPT}.  

\begin{figure*}[t!]
    \centering
    \includegraphics[width=\linewidth]{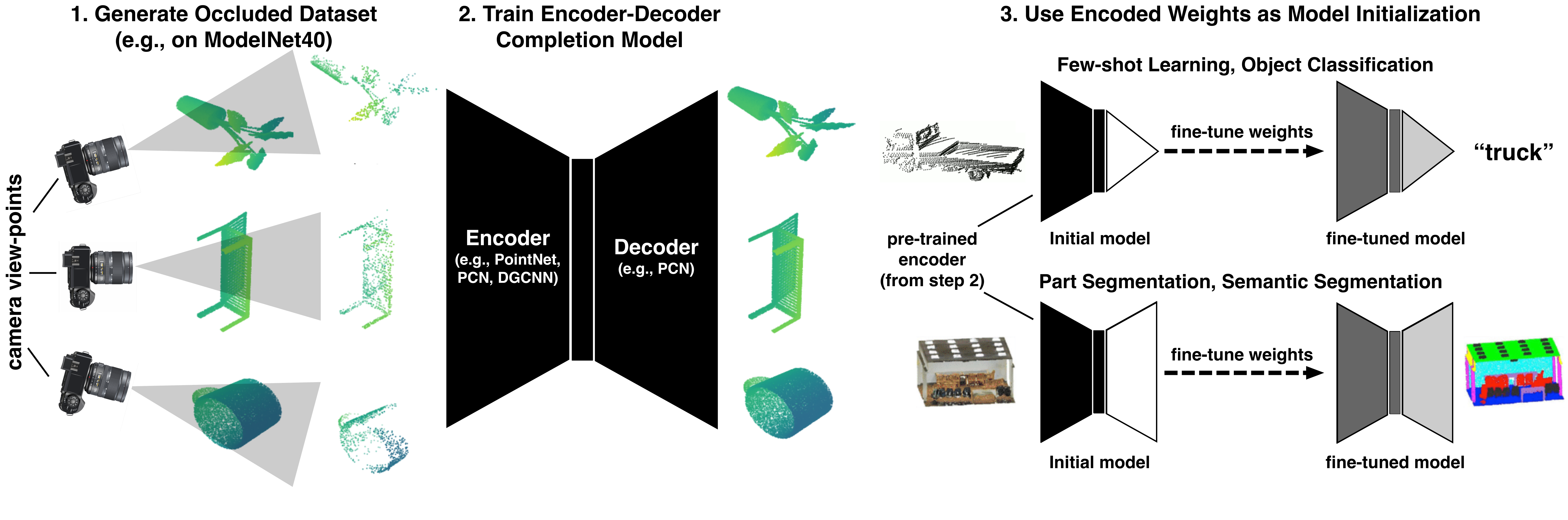}
    \vspace{-1ex}
    \caption{Overview of OcCo. 1. Take any point cloud dataset and generate occluded objects for each input by  (a) randomly sampling a camera view-point, and (b) removing points hidden from that view-point (\emph{for all experiments we use the same occluded dataset generated from ModelNet40}); 2. Train an encoder-decoder model to complete the occluded point clouds (the encoder can be any model that learns representations of point clouds, the decoder can be any completion model); 3. Use the learned encoder weights as initialisation for any downstream task (e.g., few-shot learning, object classification, part/semantic segmentation). We show that OcCo outperforms a variety of pre-training methods across multiple models and tasks.}
    \label{fig:overview}
\end{figure*}

We demonstrate that pre-training \emph{on a single object-level dataset} (ModelNet40) can improve the performance of a range of downstream tasks, even on completely different datasets. Specifically we find that OcCo has the following properties compared to other initialisation techniques: 1) \emph{Improved sample efficiency} in few-shot learning experiments; 2) \emph{Improved generalisation} in object classification, object part segmentation, and semantic segmentation; 3) \emph{Wider local minima} found after fine-tuning; 4) \emph{More semantically meaningful representations} as described via network dissection \cite{bau2017network,bau2020understanding}; 5) \emph{Better clustering quality} under jittering, translation, and rotation transformations.


    
    
    
    

\section{Related Work}\label{sec:related}


Unsupervised pre-training is gaining popularity due to its success in many problem settings, such as natural language understanding~\cite{devlin2018bert,mikolov2013efficient}, object detection~\cite{chen2020simple,he2019momentum}, graph learning~\cite{hu2019pre,GPT}, and visual localisation~\cite{schonberger2018semantic}. 
Currently, the two most common unsupervised pre-training methods for point clouds are based on
(i) generative modelling, and (ii) self-supervised learning. Work in generative modelling includes models based on generative adversarial networks (GANs)~\cite{wu2016learning,achlioptas2017learning,han2019view}, autoencoders~\cite{hassani2019unsupervised,li2018so,yang2018foldingnet}, normalizing flows~\cite{pointflow}, and approximate convex decomposition~\cite{gadelha2020label}. 

However, generative models for unsupervised pre-training on point clouds have recently been outperformed by self-supervised approaches~\cite{sauder2019self,sharma2020self,pointcontrast}. These approaches work by learning to predict key geometric properties of point clouds that are invariant across datasets. 
Specifically, \cite{sauder2019self} propose a pre-training procedure based on rearranging permuted point clouds. It works by splitting a point cloud into $k^3$ voxels, randomly permuting the voxels, and then training a model to predict the original voxel location of each point. The idea is that the pre-trained model implicitly learns about the geometric structure of point clouds by learning this rearrangement. However, there are two key issues with this objective: 1. The voxel representation is not permutation invariant. Thus, the model could learn very different representations if point clouds are rotated or translated; 2. Point clouds generated from real objects and scenes will have very different structure from randomly permuted clouds, so it is unclear why pre-trained weights that are accurate at rearrangement will be good initialisation for object classification or segmentation models. Another work \cite{sharma2020self} uses cover trees \cite{beygelzimer2006cover} to hierarchically partition points for few-shot learning. They then train a model to classify each point to their assigned partitions. However, because cover trees are designed for fast nearest neighbour search, they may arbitrarily partition semantically-contiguous regions of point clouds (e.g., airplane wings, car tires) into different regions of the hierarchy, and so ignore key point cloud geometry. 
A third work, PointContrast~\cite{pointcontrast}, uses contrastive learning to pre-train weights for point clouds of scenes. Their method uses known point-wise correspondences between different views of a complete 3D scene. These point-wise correspondences require post-processing the data by registering different depth maps into a single 3D scene. Thus, their method can only be applied to static scenes that have been registered, limiting the applicability of the approach: we leave a comparison between OcCo and PointContrast to future work. 
In what follows we will show that unsupervised pre-training based on a simple self-supervised objective: completing occluded point clouds, produces weights that outperform \cite{sauder2019self} and \cite{sharma2020self} on downstream tasks. 

Completing 3D shapes to learn model initialisations is not new,  \cite{schonberger2018semantic} used scene completion~\cite{song2017semantic,dai2020sg,hou2020revealnet} as a pre-training task to initialise 3D voxel descriptors for visual localisation. To do so, they generated partial voxelised scenes based on depth images and trained a variational autoencoder for completion. Differently, our focus is to describe a technique to learn an initialisation for point cloud models. Our aim is for this pre-trained initialisation to improve a variety of downstream tasks including few-shot learning, object classification, and segmentation, on a variety of datasets.





\section{\method}\label{sec:method}

The overall idea of our approach is shown in Figure~\ref{fig:overview}. Our observation is that by occluding point clouds based on different view-points then learning a model to complete them, the weights of the completion model can be used as initialisation for downstream tasks (e.g., classification, segmentation). This approach not only improves accuracy in few-shot learning settings but also the final generalisation accuracy in fully-supervised tasks. 

Throughout we define point clouds $\mathcal{P}$ as sets of points in 3D Euclidean space, $\mathcal{P} = \{ p_1, p_2, ..., p_n \}$, where each point $p_i$ is a vector of both coordinates $(x_i, y_i, z_i)$ and other features (e.g. colour and normal). 
We begin by describing the components that make up our occlusion mapping $o(\cdot)$. 
Then we detail how to learn a completion model $c(\cdot)$, giving pseudo-code and the architectural details in the appendix. 
 
\subsection{Generating Occlusions}
\label{subsec:occ}
We define a randomised occlusion mapping $o: \mathbb{P} \rightarrow \mathbb{P}$ (where $\mathbb{P}$ is the space of all point clouds) from a full point cloud $\mathcal{P}$ to an occluded point cloud $\tilde{\mathcal{P}}$.
This mapping constructs $\tilde{\mathcal{P}}$ by removing points from $\mathcal{P}$ that cannot be seen from a particular view-point. This is accomplished in three steps: 
(1) A projection of the complete point cloud (in a world reference frame) into the coordinates of a camera reference frame (which specifies the view-point);
(2) Identification of the points that are occluded in the camera view-point;
(3) A projection of the points back from the camera reference frame to the world reference frame. 

\paragraph{Viewing the point cloud from a camera.}
A camera defines a projection from a 3D world reference frame into a distinctive 3D camera reference frame. It does so by specifying a camera model and a camera view-point from which the projection occurs. While any camera model can be used, for illustration consider the simplest camera model: the pinhole camera. View-point projection for the pinhole camera is given by a simple linear equation:
\begin{align}
    \begin{bmatrix}
    x_{\text{cam}} \\
    y_{\text{cam}} \\
    z_{\text{cam}} \\
    \end{bmatrix} \!\!=\!\! 
    \underbrace{\vphantom{
            \left[
                \begin{array}{ccc|c}
                r_1 & r_2 & r_3 & t_1 \\
                r_4 & r_5 & r_6 & t_2 \\
                r_7 & r_8 & r_9 & t_3
                \end{array}
            \right]}
        \begin{bmatrix}
        f & \gamma & w/2 \\
        0 & f      & h/2 \\
        0 & 0      & 1
        \end{bmatrix}}_{
            \substack{\text{intrinsic} \\ [ \;\;\;\; \mathbf{K} \;\;\;\; ]}}
    \underbrace{
        \left[
        \begin{array}{ccc|c}
        r_1 & r_2 & r_3 & t_1 \\
        r_4 & r_5 & r_6 & t_2 \\
        r_7 & r_8 & r_9 & t_3
        \end{array}
        \right]
    }_{\subalign{\text{rotation } &| \text{ translation} \\ [\;\;\;\mathbf{R}\;\;\;\;\; &| \;\;\;\;\;\;\;\mathbf{t}\;\;\;\;\;]}}
    \begin{bmatrix}
    x \\
    y \\
    z \\
    1 
    \end{bmatrix}
    \label{eq:camera_transform}
\end{align}
In the above, $(x,y,z)$ are the original point cloud coordinates (in a world reference), the camera viewpoint is described by the concatenation of a rotation matrix ($r$ entries) with a translation vector ($t$ entries) describing the camera view-point, and the final matrix is the camera intrinsics ($f$ specifies the camera focal length, $\gamma$ is the skewness between the $x$ and $y$ axes in the camera, and $w,h$ are the width and height of the camera image). Given these, the final coordinates $(x_{\text{cam}},y_{\text{cam}},z_{\text{cam}})$ are the positions of the point in the camera reference frame. We will refer to the intrinsic matrix as $\mathbf{K}$ and the rotation/translation matrix as $[\mathbf{R} | \mathbf{t}]$.

\paragraph{Determining occluded points.}
We can think of the point $(x_{\text{cam}},y_{\text{cam}},z_{\text{cam}})$ in multiple ways: (a) a 3D point in the camera reference frame; (b) a 2D pixel with coordinates $(f x_{\text{cam}}/z_{\text{cam}}, f y_{\text{cam}}/z_{\text{cam}})$ with a depth of $z_{\text{cam}}$. In this way, some 2D points resulting from the projection may be occluded by others if they have the same pixel coordinates, but appear at a farther depth. To determine which points are occluded, we first use Delaunay triangulation to reconstruct a polygon mesh, then we remove the points which belong to the hidden surfaces that are determined via z-buffering \cite{strasser1974schnelle}.

\paragraph{Mapping back from camera frame to world frame.}
Once occluded points are removed, we re-project the point cloud to the original world reference frame, via the inverse transformation of eq.~\eqref{eq:camera_transform}.
Thus, the randomised occlusion mapping $o(\cdot)$ is constructed as follows. Fix an initial point cloud $\mathcal{P}$. Given a camera intrinsics matrix $\mathbf{K}$, sample rotation/translation matrices $[ [\mathbf{R}_1 | \mathbf{t}_1] , \ldots, [\mathbf{R}_V | \mathbf{t}_V]]$, where $V$ is the number of views. For each view $v \in [V]$, project $\mathcal{P}$ into the camera frame of that view-point using eq.~\eqref{eq:camera_transform}, find occluded points and remove them, then map all other points back to the world reference using its inverse. This yields the final occluded point cloud $\tilde{\mathcal{P}}_v$ for each view-point $v \in [V]$.

\subsection{The Completion Task}
Given an occluded point cloud $\tilde{\mathcal{P}}$ produced by $o(\cdot)$, the goal of the completion task is to learn a completion mapping $c: \mathbb{P} \rightarrow \mathbb{P}$ from $\tilde{\mathcal{P}}$ to a completed point cloud $\hat{\mathcal{P}}$.
A completion mapping is accurate w.r.t. loss $\ell(\cdot,\cdot)$ if $\mathbb{E}_{\tilde{\mathcal{P}} \sim o(\Pc)} \ell( c(\tilde{\mathcal{P}}), \Pc ) \rightarrow 0$.
The structure of the completion model $c(\cdot)$ is an ``encoder-decoder" network~\cite{3D-EPN,Tchapmi_2019_CVPR,WangAL20,yuan2018pcn}. The encoder maps an occluded point cloud to a vector, and the decoder completes the point cloud. After pre-training, the encoder weights can be used as initialisation for downstream tasks. In the appendix we give pseudocode for OcCo. We describe details of the completion model architecture in the following section.

\section{Experiments}\label{sec:experiments}

In this section, we present the setup of pre-training (Section \ref{subsec:pretrain_settings}) and downstream fine-tuning (Section \ref{subsec:fine_tuning_details}). Then, the results of few-shot learning, object classification, part and semantic segmentation are shown in Section \ref{subsec:fine_tuning_results}.

\subsection{OcCo Pre-Training Setup \label{subsec:pretrain_settings}}
\paragraph{Dataset.} For all experiments, we use ModelNet40~\cite{wu20153d} as the pre-training dataset. ModelNet40 includes 12,311 synthesised CAD objects from 40 categories, and the dataset is divided into 9,843/2,468 objects for training and testing, respectively. We construct a pre-training dataset using the training set. Occluded point clouds are generated with camera intrinsic parameters \{$f$=1000, $\gamma$=0, $\omega$=1600, $h$=1200\}. For each point cloud, we randomly select 10 viewpoints, where the yaw, pitch and roll angles are uniformly chosen between 0 and 2$\pi$, and the translation is set as zero. 

\paragraph{Architecture.} As described above, our pre-training
completion model $c(\cdot)$ is an encoder-decoder model. To showcase that our pre-training method is agnostic to architectures, we choose three different encoders, including PointNet~\cite{qi2017pointnet}, PCN~\cite{yuan2018pcn} and DGCNN~\cite{wang2019dynamic}. These encoders map an occluded point cloud into a 1024-dimensional vector. We adapt the folding-based decoder from~\cite{yuan2018pcn} to complete an occluded point cloud in two steps. The decoder first outputs a coarse shape consisting of 1024 points, $\hat{\mathcal{P}}_{coarse}$, then warps a 4$\times$4 2D grid around each point in $\hat{\mathcal{P}}_{coarse}$ to reconstruct a fine shape, $\hat{\mathcal{P}}_{fine}$, which consists of 16384 points. We use the Chamfer Distance (CD) as a closeness measure between prediction  $\hat{\mathcal{P}}$ and ground-truth $\mathcal{P}$:

\begin{gather}
\begin{split}
    \mathrm{CD}(\hat{\mathcal{P}}, \mathcal{P}) = &\\ \frac{1}{|\hat{\mathcal{P}}|}&\sum_{\hat{x}\in \hat{\mathcal{P}}}\min_{x\in \mathcal{P}}||\hat{x}-x||_2 + \frac{1}{|\mathcal{P}|}\sum_{x\in \mathcal{P}}\min_{\hat{x}\in \hat{\mathcal{P}}}||x-\hat{x}||_2.
\end{split}
\end{gather}
The loss of the completion model is a weighted sum of the Chamfer distances on the coarse and fine shapes:
\begin{gather}\label{eq:completion_loss}
    \ell := \mathrm{CD}(\hat{\mathcal{P}}_{coarse}, \mathcal{P}_{coarse}) + \alpha \mathrm{CD}(\hat{\mathcal{P}}_{fine}, \mathcal{P}_{fine}).
\end{gather}

\paragraph{Hyperparameters.} We use the Adam~\cite{KingmaB14} optimiser with no weight decay (L2 regularisation). The learning rate is set to 1e-4 initially and is decayed by 0.7 every 10 epochs. We pre-train the models for 50 epochs. The batch size is 32, and the momentum of batch normalisation is 0.9. The coefficient $\alpha$ in eq.~\eqref{eq:completion_loss} is set as 0.01 for the first 10000 training iterations, then increased to 0.1, 0.5 and 1.0 after 10000, 20000 and 50000 training steps, respectively.

\subsection{Fine-Tuning Setup \label{subsec:fine_tuning_details}}
\paragraph{Few-shot learning.} Few-shot learning (FSL) aims to train accurate models with very limited data. A typical setting of FSL is ``$K$-way $N$-shot''. During training, $K$ classes are randomly selected, and each category contains $N$ samples. The trained models are then evaluated on the objects from the test split. We compare OcCo with \emph{Jigsaw}~\cite{sauder2019self}, and \emph{cTree}~\cite{sharma2020self} since it outperforms previous unsupervised methods~\cite{achlioptas2017learning,wu2016learning,zhao20193d,yang2018foldingnet} as well as supervised variants~\cite{qi2017pointnet++,li2018pointcnn,qi2017pointnet,wang2019dynamic}. We follow the same setting as cTree, where we pre-train the models in a ``$K$-way $N$-shot'' configuration on ModelNet40, before evaluating on ModelNet40 and ScanObjectNN.


\paragraph{Object classification.}
Given an object represented by a set of points, object classification predicts the class that the object belongs to.
We use three 
benchmarks: ModelNet40~\cite{wu20153d}, ScanNet10~\cite{pointdan} and ScanObjectNN~\cite{ScanObjectNN}, the dataset statistics are summarised in Table~\ref{tab:cls_data}. The latter two are more challenging since they consist of occluded objects from the real-world indoor scans. We use the same settings as~\cite{qi2017pointnet,wang2019dynamic} for fine-tuning. Specifically, for PCN and PointNet, we use the Adam optimizer with an initial learning rate of 1e-3, and the learning rate is decayed by 0.7 every 20 epochs with the minimum value 1e-5. For DGCNN, we use the SGD optimizer with momentum 0.9 and weight decay 1e-4. The learning rate starts from 0.1 and then decays using cosine annealing~\cite{LoshchilovH17} with  the minimum value 1e-3. We use dropout~\cite{dropout} in the fully connected layers before the softmax output layer. The dropout rate is set to 0.7 for PointNet and PCN and is set to 0.5 for DGCNN. For all three models, we train them for 200 epochs with batch size 32. We report the test results based on three runs in Table~\ref{tab:classification}.
\begin{table}[t!]
\centering
\setlength{\tabcolsep}{3pt}
\caption{Statistics of classification datasets}
\begin{tabular}{l|c|c|c}
\bottomrule 
Name & Type & \# Class & \# Training/Testing  \\\toprule \bottomrule
ModelNet & synthesised & 40 & 9,843\,/\,2,468 \\\hline
ScanNet & real scanned & 10 & 6,110\,/\,1,769 \\\hline
ScanObjectNN & real scanned & 15 & 2,304\,/\,576 \\
\toprule
\end{tabular}
\label{tab:cls_data}
\vspace{-2ex}
\end{table}
\paragraph{Part segmentation.} 
Part segmentation is a challenging fine-grained 3D recognition task. The mission is to predict the part category label (e.g., chair leg, cup handle) of each point for a given object.
To evaluate the effectiveness of OcCo pre-training, we use ShapeNetPart~\cite{armeni20163d} benchmark, which contains 16,881 objects from 16 categories and has 50 parts in total. Each object is represented by 2048 points. For PCN and PointNet, we use the Adam optimizer with an initial learning rate of 1e-3, and the learning rate is decayed by 0.5 every 20 epochs with the minimum value 1e-5. For DGCNN, we use an SGD optimizer with momentum 0.9 and weight decay 1e-4. The learning rate starts from 0.1 and then decays using cosine annealing~\cite{LoshchilovH17} with the minimum value 1e-3. We train the models for 250 epochs with batch size 16. We use the same post-processing during testing as~\cite{qi2017pointnet} and report the results over three runs in Table~\ref{tab:partseg_result}.

\paragraph{Semantic segmentation.} 
Semantic segmentation predicts the semantic object category of each point under an indoor/outdoor scene. We use S3DIS benchmark~\cite{armeni20163d} for indoor scene segmentation and SensatUrban benchmark~\cite{hu2020towards} for outdoor scene segmentation. S3DIS contains 3D scans collected via Matterport scanners in 6 different places, encompassing 271 rooms and 13 semantic classes. While SensatUrban consists of over three billion annotated points, covering large areas in a total of 7.6 km$^2$ from three UK cities (Birmingham, Cambridge, and York). Each point in SensatUrban is labelled as one of 13 semantic classes.
We use the same pre-processing, post-processing and training settings as ~\cite{qi2017pointnet,wang2019dynamic}. Each point is described by a 9-dimensional vector (coordinates, RGBs and normalised location). We train all the models for 100 epochs with batch size 24. We report the results based on three runs in Table~\ref{tab:seg_result}.
\begin{table}[t!]
\centering
\setlength{\tabcolsep}{2pt}
\caption{Few-shot learning results. We report mean and standard error over 10 runs and bold the best results.\label{tab:fsl}}
\begin{tabular}{l|cccc}
\bottomrule \multicolumn{1}{c|}{\multirow{2}{*}{Baseline}}  & 
\multicolumn{2}{c}{5-way} & \multicolumn{2}{c}{10-way} \\\cline{2-5}
& 10-shot & 20-shot & 10-shot & 20-shot \\ \toprule \bottomrule
& \multicolumn{4}{c}{ModelNet40}\\\hline
PointNet, Rand & 52.0$\pm$3.8 & 57.8$\pm$4.9 & 46.6$\pm$4.3 & 35.2$\pm$4.8 \\
PointNet, Jigsaw & 66.5$\pm$2.5 & 69.2$\pm$2.4 & 56.9$\pm$2.5 & 66.5$\pm$1.4\\
PointNet, cTree  & 63.2$\pm$3.4 & 68.9$\pm$3.0 & 49.2$\pm$1.9 & 50.1$\pm$1.6 \\
PointNet, OcCo & \textcolor{blue}{89.7$\pm$1.9} & \textcolor{blue}{92.4$\pm$1.6} & \textbf{83.9$\pm$1.8} & \textbf{89.7$\pm$1.5} \\\hline
DGCNN, Rand   & 31.6$\pm$2.8 & 40.8$\pm$4.6 & 19.9$\pm$2.1 & 16.9$\pm$1.5\\
DGCNN, Jigsaw & 34.3$\pm$1.3 & 42.2$\pm$3.5 & 26.0$\pm$2.4 & 29.9$\pm$2.6\\
DGCNN, cTree  & 60.0$\pm$2.8 & 65.7$\pm$2.6 & 48.5$\pm$1.8 & 53.0$\pm$1.3\\
DGCNN, OcCo & \textbf{90.6$\pm$2.8} & \textbf{92.5$\pm$1.9} & \textcolor{blue}{82.9$\pm$1.3} & \textcolor{blue}{86.5$\pm$2.2} \\
\toprule \bottomrule
& \multicolumn{4}{c}{ScanObjectNN}\\\hline
PointNet, Rand   & 57.6$\pm$2.5 & 61.4$\pm$2.4 & 41.3$\pm$1.3 & 43.8$\pm$1.9 \\
PointNet, Jigsaw & 58.6$\pm$1.9 & 67.6$\pm$2.1 & 53.6$\pm$1.7 & 48.1$\pm$1.9 \\
PointNet, cTree  & 59.6$\pm$2.3 & 61.4$\pm$1.4 & 53.0$\pm$1.9 & 50.9$\pm$2.1 \\
PointNet, OcCo & \textcolor{blue}{70.4$\pm$3.3} & \textcolor{blue}{72.2$\pm$3.0} & \textcolor{blue}{54.8$\pm$1.3} & \textbf{61.8$\pm$1.2} \\\hline
DGCNN, Rand   & 62.0$\pm$5.6 & 67.8$\pm$5.1 & 37.8$\pm$4.3 & 41.8$\pm$2.4 \\
DGCNN, Jigsaw & 65.2$\pm$3.8 & 72.2$\pm$2.7 & 45.6$\pm$3.1 & 48.2$\pm$2.8 \\
DGCNN, cTree  & 68.4$\pm$3.4 & 71.6$\pm$2.9 & 42.4$\pm$2.7 & 43.0$\pm$3.0 \\
DGCNN, OcCo & \textbf{72.4$\pm$1.4} & \textbf{77.2$\pm$1.4} & \textbf{57.0$\pm$1.3} & \textcolor{blue}{61.6$\pm$1.2} \\\hline
\toprule
\end{tabular}
\end{table}
\begin{table*}[!ht]
\centering
\setlength{\tabcolsep}{4.5pt}
\caption{Overal accuracy on 3D object classification benchmarks. We reported the mean and standard error over three runs.}
\vspace{-1ex}
\begin{tabular}{l|c|c|c|c|c|c|c|c|c}
\bottomrule \multirow{2}{*}{Dataset} & 
\multicolumn{3}{c|}{PointNet} & \multicolumn{3}{c|}{PCN} & \multicolumn{3}{c}{DGCNN} \\ \cline{2-10}
 & Random & Jigsaw & \abbrev& Random & Jigsaw & \abbrev & Random & Jigsaw & \abbrev \\ \toprule \bottomrule
ModelNet& 89.2$\pm$0.1 & 89.6$\pm$0.1 & \textcolor{blue}{90.1$\pm$0.1} & 89.3$\pm$0.1 & 89.6$\pm$0.2 & \textcolor{blue}{90.3$\pm$0.2} & 92.5$\pm$0.4 & 92.3$\pm$0.3 & \textbf{93.0$\pm$0.2} \\\hline
ScanNet& 76.9$\pm$0.2 & 77.2$\pm$0.2 & \textcolor{blue}{78.0$\pm$0.2} & 77.0$\pm$0.3 & 77.9$\pm$0.3 & \textcolor{blue}{78.2$\pm$0.3} & 76.1$\pm$0.7 & 77.8$\pm$0.5 & \textbf{78.5$\pm$0.3}  \\\hline
ScanObjectNN & 73.5$\pm$0.5 & 76.5$\pm$0.4 & \textcolor{blue}{80.0$\pm$0.2} & 78.3$\pm$0.3 & 78.2$\pm$0.1 & \textcolor{blue}{80.4$\pm$0.2} & 82.4$\pm$0.4 & 82.7$\pm$0.8 & \textbf{83.9$\pm$0.4} \\
\toprule
\end{tabular}
\label{tab:classification}
\end{table*}
\begin{table*}[t!]
\setlength{\tabcolsep}{5.5pt}
\centering
\vspace{-1ex}
\caption{Overall accuracy and intersection of union (mIoU) on ShapeNetPart. We reported the mean and ste over three runs.}
\vspace{-1ex}
\begin{tabular}{l|c|c|c|c|c|c|c|c|c}
\bottomrule \multirow{2}{*}{} & 
\multicolumn{3}{c|}{PointNet} & \multicolumn{3}{c|}{PCN} & \multicolumn{3}{c}{DGCNN} \\ \cline{2-10}
 & Random & Jigsaw & \abbrev& Random & Jigsaw & \abbrev & Random & Jigsaw & \abbrev \\ \toprule \bottomrule
OA (\%) & 92.8$\pm$0.9 & 93.1$\pm$0.5 & \textcolor{blue}{93.4$\pm$0.7} & 92.3$\pm$1.0 & 92.6$\pm$0.9 & \textcolor{blue}{93.0$\pm$0.9} & 92.2$\pm$0.9 & 92.7$\pm$0.9 & \textbf{94.4$\pm$0.7} \\\hline
mIoU (\%) & 82.2$\pm$2.4 & 82.2$\pm$2.8 & \textcolor{blue}{83.4$\pm$1.9} & 81.3$\pm$2.6 & 81.2$\pm$2.9 & \textcolor{blue}{82.3$\pm$2.4} & 84.4$\pm$1.2 & 84.3$\pm$1.2 & \textbf{85.0$\pm$1.0} \\
\toprule
\end{tabular}
\label{tab:partseg_result}
\end{table*}
\begin{table*}[t!]
\setlength{\tabcolsep}{5.5pt}
\centering
\caption{Overall accuracy (OA) and mean intersection of union (mIoU) on the S3DIS across six folds over three runs.}
\vspace{-1ex}
\begin{tabular}{l|c|c|c|c|c|c|c|c|c}
\bottomrule \multirow{2}{*}{} & 
\multicolumn{3}{c|}{PointNet} & \multicolumn{3}{c|}{PCN} & \multicolumn{3}{c}{DGCNN} \\ \cline{2-10}
 & Rand & Jigsaw & \abbrev& Rand & Jigsaw & \abbrev & Rand & Jigsaw & \abbrev \\ \toprule \bottomrule
OA (\%) & 78.2$\pm$0.7 & 80.1$\pm$1.2 & \textcolor{blue}{82.0$\pm$1.0} & 82.9$\pm$0.9 & 83.7$\pm$0.7 & \textbf{85.1$\pm$0.5} & 83.7$\pm$0.7 & 84.1$\pm$0.7 & \textcolor{blue}{84.6$\pm$0.5} \\\hline
mIoU (\%) & 47.0$\pm$1.4 & 52.6$\pm$1.9 & \textcolor{blue}{54.9$\pm$1.0} & 51.1$\pm$2.4 & 52.2$\pm$1.9 & \textcolor{blue}{53.4$\pm$2.1} & 54.9$\pm$2.1 & 55.6$\pm$1.4 & \textbf{58.0$\pm$1.7}\\
\toprule
\end{tabular}
\label{tab:seg_result}
\end{table*}
\begin{table*}[t!]\small
\setlength{\tabcolsep}{3.5pt}
\centering
\caption{Overall point accuracy (OA), mean class accuracy (mAcc) and mean class intersection of union (mIoU) on SensatUrban. We reported the mean and standard error over three runs. We use the same preprocess procedures as PointNet.}
\vspace{-1ex}
\begin{tabular}{ccccccccccccccccc}\bottomrule 
& \rotatebox{90}{OA(\%)} & \rotatebox{90}{mAcc(\%)} & \rotatebox{90}{mIoU(\%)\,\,} & \rotatebox{90}{ground} & \rotatebox{90}{veg} & \rotatebox{90}{building} & \rotatebox{90}{wall} & \rotatebox{90}{bridge} & \rotatebox{90}{parking} & \rotatebox{90}{rail} & \rotatebox{90}{traffic} & \rotatebox{90}{street} & \rotatebox{90}{car} & \rotatebox{90}{footpath} & \rotatebox{90}{bike} & \rotatebox{90}{water} \\\toprule
PointNet & 86.29 & 53.33 & 45.10 & 80.05 & 93.98 & 87.05 & 23.05 & 19.52 & 41.80 & 3.38 & 43.47 & 24.20 & 63.43 & 26.86 & 0.00 & 79.53 \\ 
PointNet-Jigsaw & 87.38 & 56.97 & 47.90 & 83.36 & 94.72 & 88.48 & 22.87 & 30.19 & 47.43 & 15.62 & 44.49 & 22.91 & 64.14 & 30.33 & 0.00 & 77.88 \\ 
PointNet-OcCo & 87.87 & 56.14 & 48.50 & 83.76 & 94.81 & 89.24 & 23.29 & 33.38 & 48.04 & 15.84 & 45.38 & 24.99 & 65.00 & 27.13 & 0.00 & 79.58 \\\toprule\bottomrule
PCN & 86.79 & 57.66 & 47.91 & 82.61 & 94.82 & 89.04 & 26.66 & 21.96 & 34.96 & 28.39 & 43.32 & 27.13 & 62.97 & 30.87 & 0.00 & 80.06 \\
PCN-Jigsaw & 87.32 & 57.01 & 48.44 & 83.20 & 94.79 & 89.25 & 25.89 & 19.69 & 40.90 & 28.52 & 43.46 & 24.78 & 63.08 & 31.74 & 0.00 & 84.42 \\
PCN-OcCo & 86.90 & 58.15 & 48.54 & 81.64 & 94.37 & 88.21 & 25.43 & 31.54 & 39.39 & 22.02 & 45.47 & 27.60 & 65.33 & 32.07 & 0.00 & 77.99 \\\toprule\bottomrule
DGCNN  & 87.54 & 60.27 & 51.96 & 83.12 & 95.43 & 89.58 & \textbf{31.84} & 35.49 & 45.11 & 38.57 & 45.66 & 32.97 & 64.88 & 30.48 & 0.00 & \textbf{82.34} \\
DGCNN-Jigsaw  & 88.65 & 60.80 & 53.01 & \textbf{83.95} & \textbf{95.92} & 89.85 & 30.05 & \textbf{43.59} & 46.40 & 35.28 & 49.60 & 31.46 & 69.41 & \textbf{34.38} & 0.00 & 80.55\\
DGCNN-OcCo & \textbf{88.67} & \textbf{61.35} & \textbf{53.31} & 83.64 & 95.75 & \textbf{89.96} & 29.22 & 41.47 & \textbf{46.89} & \textbf{40.64} & \textbf{49.72} & \textbf{33.57} & \textbf{70.11} & 32.35 & 0.00 & 79.74 \\ \toprule
\end{tabular}
\label{tab:outseg_result}
\end{table*}

\subsection{Fine-Tuning Results \label{subsec:fine_tuning_results}}
\paragraph{Few-shot learning.} 
We report the experimental results on few-shot learning in Table~\ref{tab:fsl}. We colour the best results with \textcolor{blue}{blue} for each encoder and \textbf{bold} the overall best score for each dataset. We use the same colouring scheme in all subsequent results. We find that \abbrev outperforms both few-shot baselines Jigsaw~\cite{sauder2019self} and cTree~\cite{sharma2020self} in-domain (ModelNet40) and cross-domain (ScanObjectNN). 
We believe this is due to the fact that the occlusions \abbrev generates will be due to the geometric structure of the object, whereas the voxel permutations of \cite{sauder2019self} and the cover tree partitioning of \cite{sharma2020self} may destroy aspects of this structure.

\paragraph{Object classification.} 
Table~\ref{tab:classification} compares OcCo with random and Jigsaw~\cite{sauder2019self} initialisation on object classification.\footnote{Note we intentionally did not compare with cTree~\cite{sharma2020self} as it is specifically designed for few-shot learning.} We show that OcCo-initialised models outperform these baselines on all datasets. OcCo performs well not only on the in-domain dataset (ModelNet), but also on cross-domain datasets (ScanNet and ScanObjectNN). The improvements are consistent across the three encoders. 
In the following section we will provide one explanation: the local minima found after fine-tuning an OcCo-based initialisation appear to be wider than those found using other initialisations.

\begin{figure*}[ht!]
    \centering
    \includegraphics[width=1\textwidth]{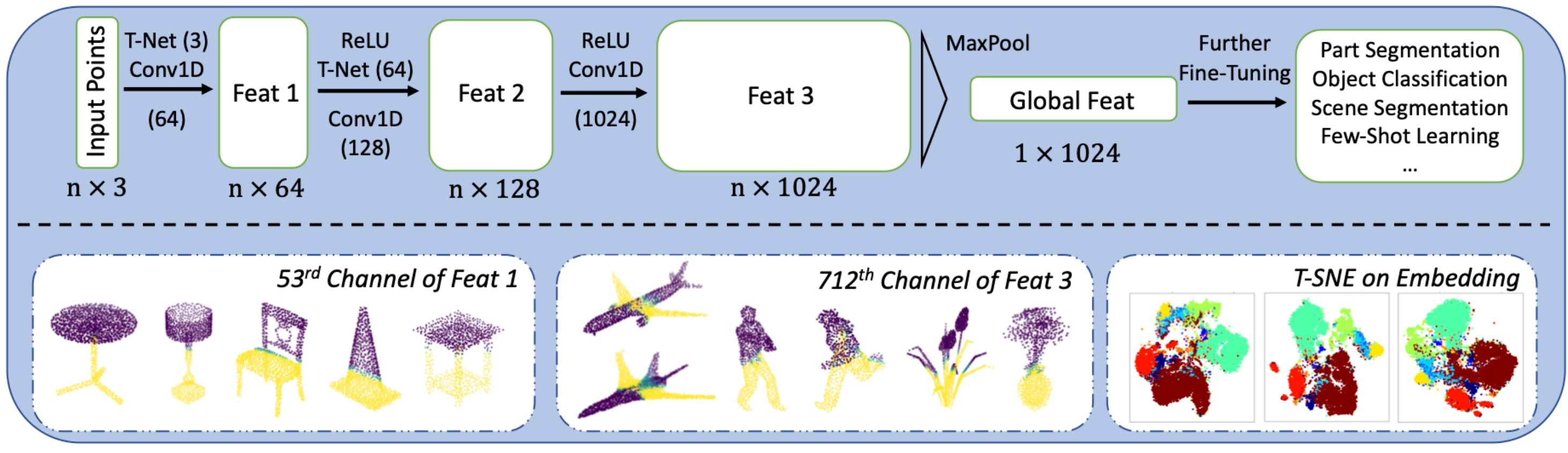}
    \caption{Visualisation on the learned features by OcCo-PointNet.}
    \label{fig:tsne}
\vspace{2ex}
\end{figure*}
\paragraph{Object part segmentation.}
Table~\ref{tab:partseg_result} compares OcCo-initialisation with random and Jigsaw~\cite{sauder2019self} initialisation on object part segmentation. We observe that OcCo-initialised models outperform the others in terms of overall accuracy and mean class IoU. These results are consistent across various encoders. 
We further analyse why OcCo helps the encoders better recognise the object parts with feature visualisation and concept detection in Section~\ref{sec:analysis}.

\paragraph{Semantic segmentation.} 
We compare random, Jigsaw and \abbrev initialisation on both indoor and outdoor semantic segmentation tasks. For S3DIS, we evaluate the trained models using $6$-fold cross-validation following~\cite{armeni20163d}, and report the scores in Table~\ref{tab:seg_result}. It is clear that OcCo-initialised models outperform random and Jigsaw-initialised ones. For SensatUrban, we report the scores in Table~\ref{tab:outseg_result}. We observe that OcCo outperforms random initialisation and Jigsaw initialisation for semantic categories that are included in the pre-training dataset, such as cars. For classes that are not included in ModelNet40, OcCo is competitive with the other methods. This makes sense as the geometries of these objects are likely not well understood by the learned initialisations. Ultimately, we find it encouraging that OcCo which learns representations at the object-level can still improve generalisation on segmentation on outdoor scenes.



\section{Analysis}\label{sec:analysis}
In this section, we first show that \abbrev pre-training leads to a fine-tuned model that converges to a local minimum that is flatter than other initialisations. Then we evaluate the learned representations from \abbrev with feature visualisation, semantic concept detection and unsupervised mutual information. The analysis demonstrates that \abbrev can learn rich and discriminative point cloud features.

\begin{figure}[h!]
    \centering
     \includegraphics[width=\linewidth]{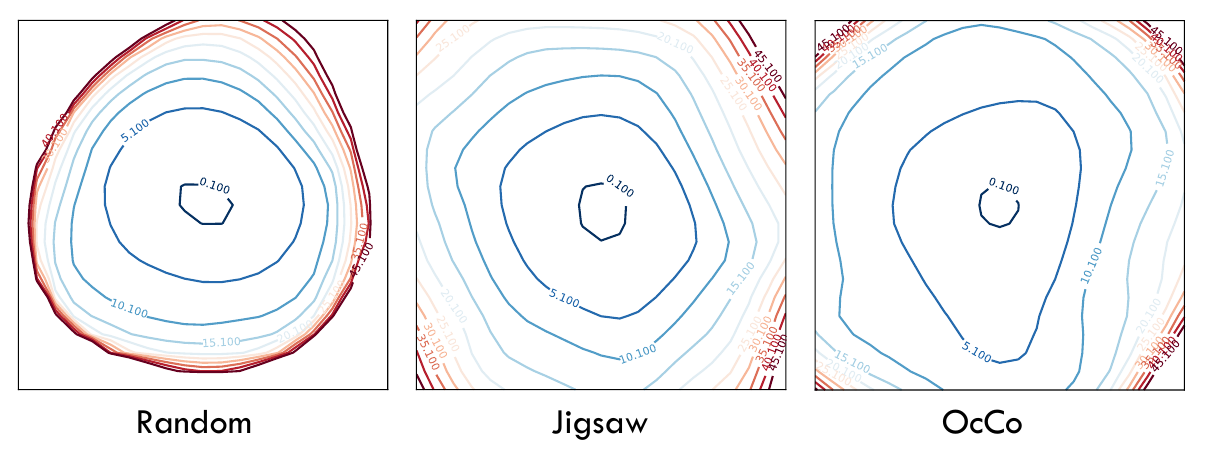}
     \vspace{-2ex}
     \caption{Loss landscape visualisation.\label{fig:losslandscape}}
     \vspace{-1ex}
\end{figure}

\paragraph{Visualisation of optimisation landscape.}
We follow the same procedure of~\cite{Li0TSG18} to visualise the loss landscapes of random, Jigsaw and OcCo initialised PointNet in Figure~\ref{fig:losslandscape}. All three models are fine-tuned on ScanObjectNN with the training settings described in Section~\ref{subsec:fine_tuning_details}. For visualisation, we use two random vectors, $\boldsymbol{\delta}$ and $\boldsymbol{\eta}$, to perturb the fine-tuned parameters $\boldsymbol{\theta}^{*}$ and obtain corresponding loss values. The 2D plot $f(\alpha, \beta)$ is defined as:
\begin{gather}
    f(\alpha, \beta)=\mathcal{L}\left(\boldsymbol{\theta}^{*}+\alpha \boldsymbol{\delta}+\beta \boldsymbol{\eta}\right)
\end{gather}
where each filter in $\boldsymbol{\delta}$ and $\boldsymbol{\eta}$ is normalised w.r.t the corresponding filter in $\boldsymbol{\theta}^{*}$. $\alpha$ and $\beta$ have the same ranges of $[-1, 1]$. 
We observe that the model with OcCo pre-training can converge to a flatter local minimum,
which is known to have better generalisation~\cite{ChaudhariCSLBBC17,hochreiter1997flat}.
\begin{figure*}[ht!]
    \centering
    \includegraphics[width=\textwidth]{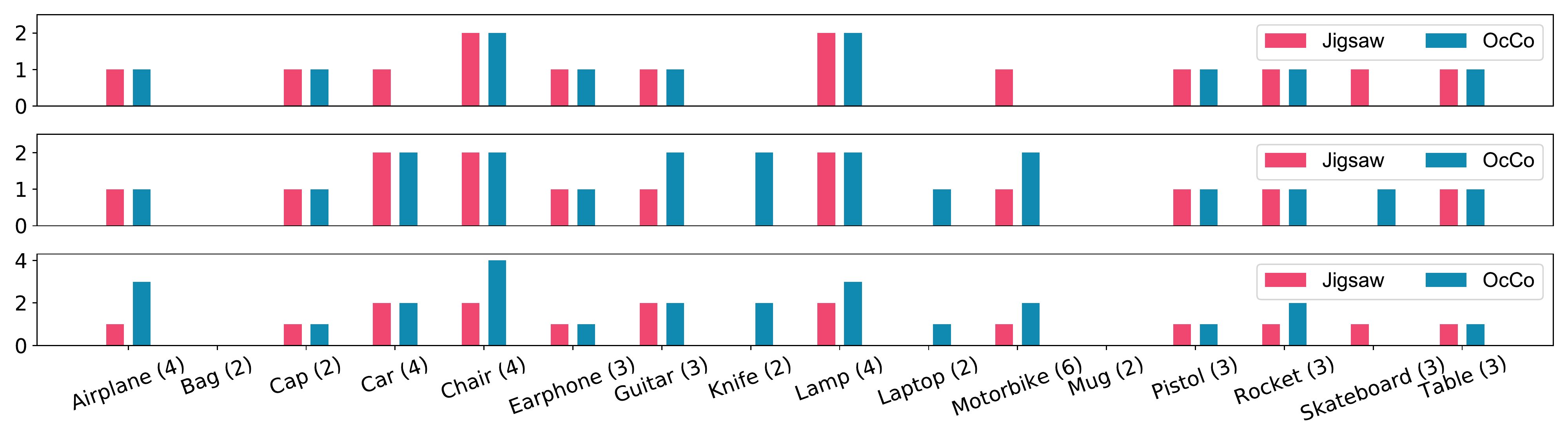}
    \caption{Number of detected object parts in the `Feat1'~(above), `Feat2'~(middle) and `Feat3'~(below) module of Jigsaw and OcCo-initialised PointNet feature encoder. Digits in the brackets are the number of parts under that object category.}
    \label{fig:feat-det}
    \vspace{1.5ex}
\end{figure*}
\begin{table*}[ht!]\small
\setlength{\tabcolsep}{6.8pt}
\centering
\caption{Adjusted mutual information (AMI) under transformations. We report the mean and standard error over 10 random initialisation. Under the `transformation' column, `J', `T', `R' represent jittering, translation and rotation, respectively.}
\vspace{-1ex}
\begin{tabular}{c|c|c|c|c|c|c|c|c|c|c}
\bottomrule \multicolumn{3}{c|}{Transformation} & \multicolumn{4}{c|}{ShapeNet10} & \multicolumn{4}{c}{ScanObjectNN} \\ \hline
J & T & R & VFH & M2DP & Jigsaw & \abbrev & VFH & M2DP & Jigsaw & \abbrev \\ \toprule \bottomrule
& & & 0.12$\pm$0.01 & 0.22$\pm$0.03 & 0.33$\pm$0.04 & \textbf{0.51$\pm$0.03} & 0.05$\pm$0.02 & 0.18$\pm$0.02 & 0.29$\pm$0.02 & \textbf{0.44$\pm$0.03} \\\hline
\checkmark & & & 0.12$\pm$0.02 & 0.19$\pm$0.02 & 0.32$\pm$0.02 & \textbf{0.45$\pm$0.02} & 0.06$\pm$0.02 & 0.17$\pm$0.02 & 0.27$\pm$0.02 & \textbf{0.42$\pm$0.04} \\\hline
\checkmark & \checkmark & & 0.13$\pm$0.03 & 0.21$\pm$0.02 & 0.29$\pm$0.07 & \textbf{0.38$\pm$0.04} & 0.04$\pm$0.02 & 0.18$\pm$0.03 & 0.24$\pm$0.04 & \textbf{0.39$\pm$0.06} \\\hline
\checkmark & \checkmark & \checkmark & 0.07$\pm$0.03 & 0.20$\pm$0.04 & 0.28$\pm$0.03 & \textbf{0.35$\pm$0.05} & 0.04$\pm$0.01 & 0.16$\pm$0.03 & 0.18$\pm$0.09 & \textbf{0.34$\pm$0.06} \\\hline
\toprule
\end{tabular}
\label{tab:so3}
\end{table*}
\paragraph{Visualisation of learned features.} 
We use feature visualisation to explore what a pre-trained model has learned about point cloud objects before fine-tuning. In Figure~\ref{fig:tsne}, we visualise the features/embeddings of the objects from the test split of ModelNet40. 
We colour the points according to their channel activations. The larger the activation value is, the darker the colour will be. 
We observe that the pre-trained encoder can learn low-level geometric primitives, e.g., planes, cylinders and cones, in the early stage. While it later recognises more complex shapes like wings, leaves and upper bodies. We further use t-SNE to visualise the object embeddings on ShapeNet10. We notice that distinguishable clusters are formed after pre-training. Thus, it seems that \abbrev can learn features that are useful to distinguish different parts of an object or a scene. These features will be beneficial to downstream tasks, e.g., object classification and scene segmentation.
\paragraph{Unsupervised mutual information probe.} 
We hypothesise that a pre-trained model without fine-tuning can learn label information in an unsupervised fashion, i.e., zero-shot learning on cross-domain datasets. To validate, we utilise OcCo-PointNet to extract global features for objects from ShapeNet10 and ScanObjectNN. Then, we cluster the extracted embeddings with an unsupervised clustering method, K-means (where K is set to the number of object categories). To evaluate the clustering quality, we calculate the adjusted mutual information (AMI)~\cite{nguyen2009information} between the generated and the ground-truth clusters. AMI reaches 1 if two clusters are identical, while it has an expected value of 0 for a random categorical cluster assignment. Besides, we also study whether the OcCo-PointNet is robust to input transformations. In particular, we consider three transformations, including rotation, translation and jittering. We apply these transformations to an input point cloud before using PointNet for feature/embedding extraction. 

We compare \abbrev with Jigsaw and two hand-crafted point cloud global descriptors: viewpoint feature histogram (VFH)~\cite{rusu2010fast} and M2DP~\cite{he2016m2dp} in Table~\ref{tab:so3}. We observe that pre-training methods, e.g., Jigsaw and \abbrev, can learn more discriminative feature representations than hand-crafted descriptors, while the representations learned from \abbrev pre-trained encoder are more predictive than Jigsaw based method. These results demonstrate that \abbrev is effective for unsupervised feature learning.

\begin{figure}[h!]
    \centering
     \includegraphics[width=\linewidth]{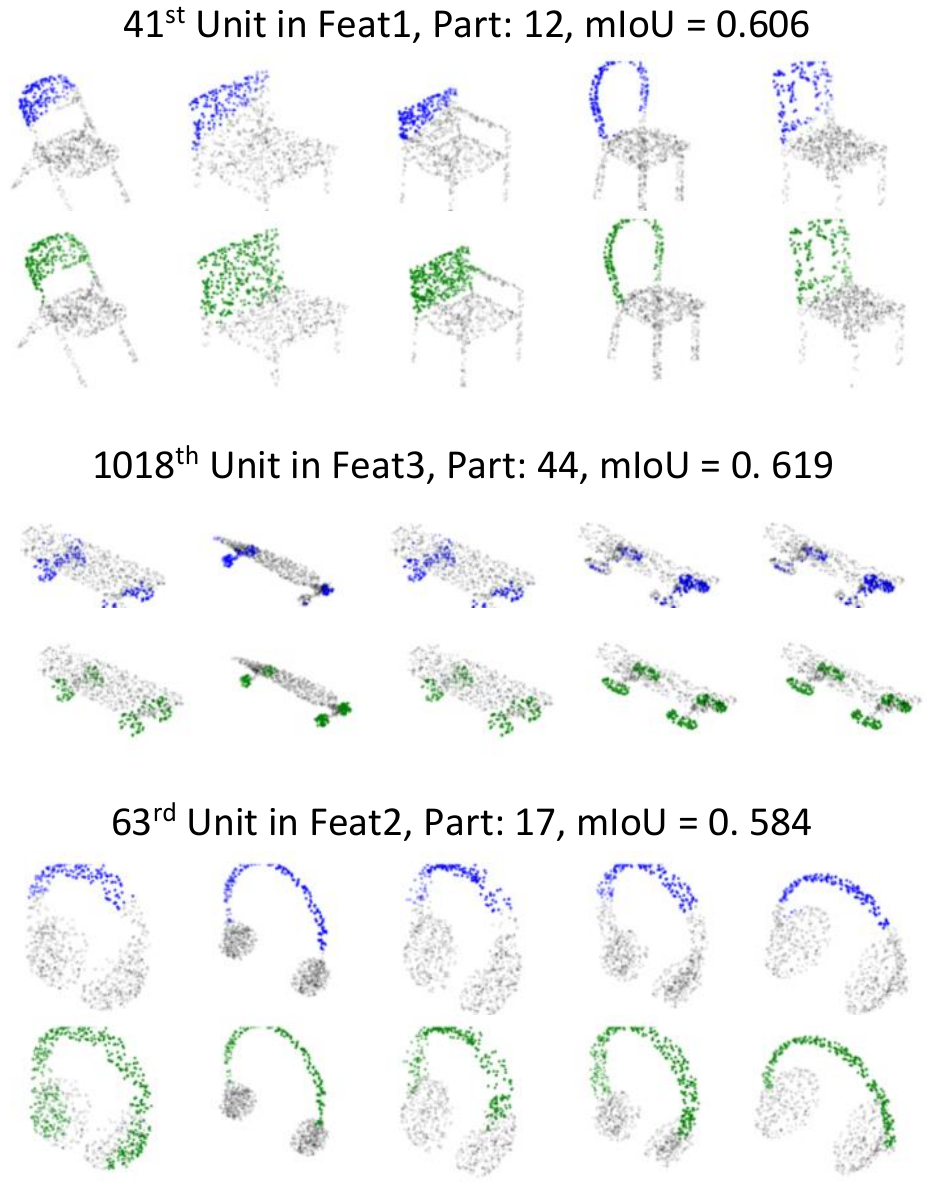}
     \caption{Visualisation of detected concepts. Parts marked by blue and green are the binary masks based on the feature activations ($M_k$) and the ground truth labels ($C_n$).\label{fig:point-dissection}}
\end{figure}
\paragraph{Detection of semantic concepts.} 
We adapt \emph{network dissection}~\cite{bau2017network,bau2020understanding} to study whether OcCo pre-trained models can learn semantic concepts in an unsupervised fashion without fine-tuning. Specifically, for each object, we first create an activation mask $M_k$ based on the feature map from the $k$-th channel in the network. We assign the $i$-th entry of $M_k$ as 1 if the activation of the $i$-th point in that feature map is among the top 20\%, otherwise the $i$-th entry is assigned to 0. The concept mask $C_n$ marks the points as 1 if they belong to the $n$-th semantic concept (e.g., chair legs) in the ground truth annotations. Given a set of point clouds~$\mathcal{D}_\mathcal{P}$, we calculate the mean intersection of union (mIoU) scores based on these binary masks:
\begin{gather}
    \text{mIoU}_{(k, n)}=
    \mathbb{E}_{\mathcal{P}\sim\mathcal{D}_\mathcal{P}}
    \left[\frac{\left|M_{k}(\mathcal{P}) \cap C_{n}(\mathcal{P})\right|}
    {\left|M_{k}(\mathcal{P}) \cup C_{n}(\mathcal{P})\right|}\right]
\end{gather}
where $|\cdot|$ is the set cardinality. 
$\text{mIoU}_{(k, n)}$ can be interpreted as how well channel $k$ detects the concept $n$. In Figure~\ref{fig:feat-det}, we plot the number of detected concepts (i.e., $\text{mIoU}_{(k, n)}>0.5$). We conclude that \abbrev outperforms Jigsaw in terms of the total number of detected concepts. We visualise some masks from OcCo-PointNet in Figure~\ref{fig:point-dissection}. We observe that OcCo pre-training can capture rich concept information. These results demonstrate that pre-training with \abbrev can unsupervisedly learn semantic concepts.

\section{Discussion}\label{sec:discussion}

In this work, we have demonstrated that Occlusion Completion (OcCo) can learn representations for point clouds that are accurate in few-shot learning, in object classification, and in part and semantic segmentation tasks, as compared to prior work. We performed multiple analyses to explain why this occurs, including a visualisation of the loss landscape, visualisation of learned features, tests of transformation invariance, and quantifying how well the initialisations can learn semantic concepts. In the future, it would be interesting to design a completion model that is explicitly aware of the occlusion procedure. 
This model would may converge even quicker and require fewer parameters, as this could act as a stronger inductive bias during learning. 

\section{Acknowledgements}
We would like to thank Qingyong Hu, Shengyu Huang, Matthias Niessner, Kilian Q. Weinberger, and Trevor Darrell for valuable discussions and feedbacks.

\newpage
{\small
\bibliographystyle{ieee_fullname}
\bibliography{egbib}
}

\newpage
\appendix
\onecolumn

\section{Implementation details}
\paragraph{Completion pre-training}
Previous point completion models \cite{3D-EPN, yuan2018pcn,Tchapmi_2019_CVPR,WangAL20} all use an "encoder-decoder" architecture. The encoder maps a partial point cloud to a vector of a fixed dimension, and the decoder reconstructs the full shape. 

\begin{figure}[H]
    \centering
     \includegraphics[width=\textwidth]{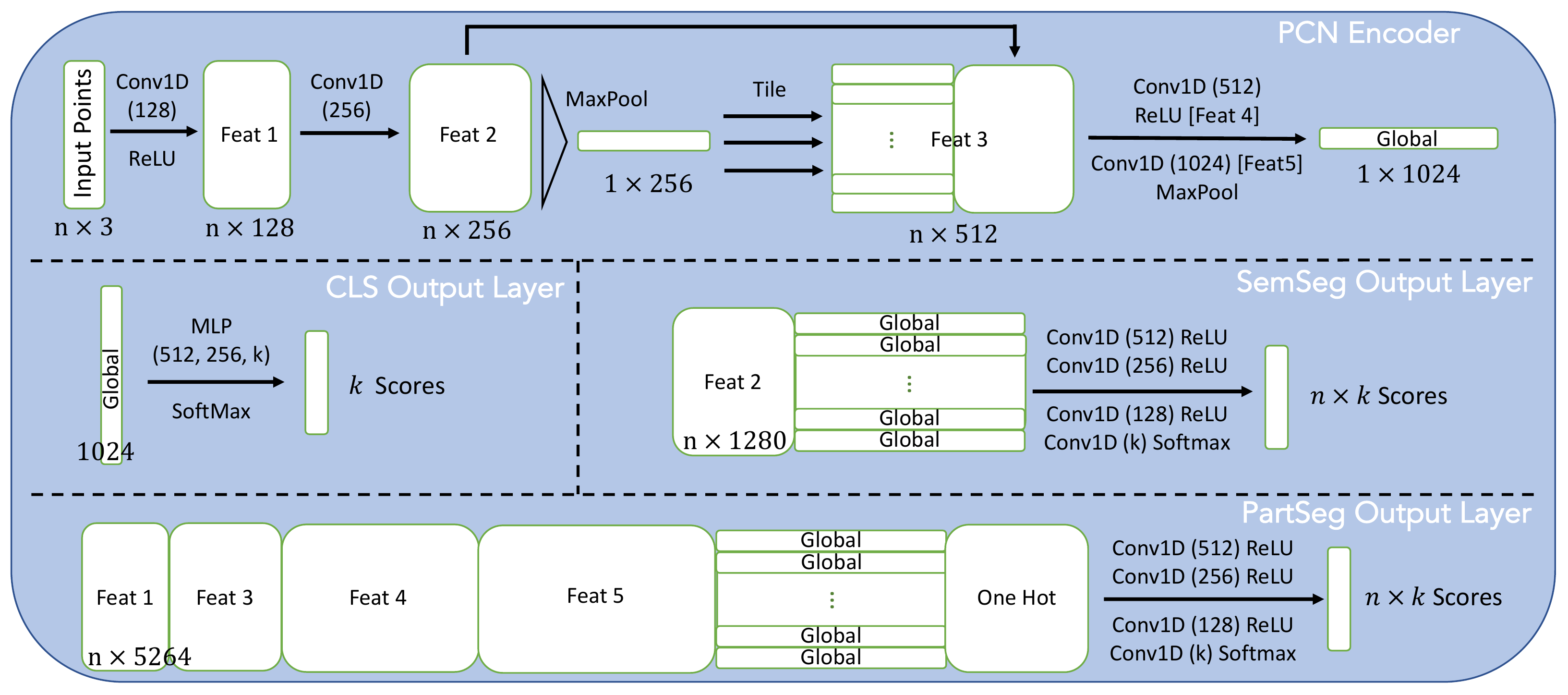}
     \caption{Encoder and Output Layers of PCN}
     \label{fig:pppcn}
\end{figure}

In the \abbrev experiments, we exclude the last few MLPs of PointNet and DGCNN, and use the remaining architecture as the encoder to map a partial point cloud into a 1024-dimensional vector. 
We adapt the folding-based decoder design from PCN, which is a two-stage point cloud generator that generates a coarse and a fine-grained output point cloud $(Y_{coarse}, Y_{fine})$ for each input feature. We sketch the network structures of PCN encoder and output layers for downstream tasks in Figure~\ref{fig:pppcn}. We removed all the batch normalisation in the folding-based decoder since we find they bring negative effects in the completion process in terms of loss and convergence rate, this has been reported in image generations~\cite{park2019semantic}.
Also, we find L2 normalisation in the Adam optimiser is undesirable for completion training but brings improvements for the downstream fine-tuning tasks.


We compare the occluded datasets based on ModelNet40 and ShapeNet8 for the \abbrev pre-training. We construct the ModelNet Occluded using the methods described in Section~\ref{sec:method} and for ShapeNet Occluded we directly use the data provided in the PCN, whose generation method are similar but not exactly the same with ours. Basic statistics of these two datasets are reported in Table~\ref{tab:occo_stats}. 

\begin{figure}[H]
    \centering
     \includegraphics[width=\textwidth]{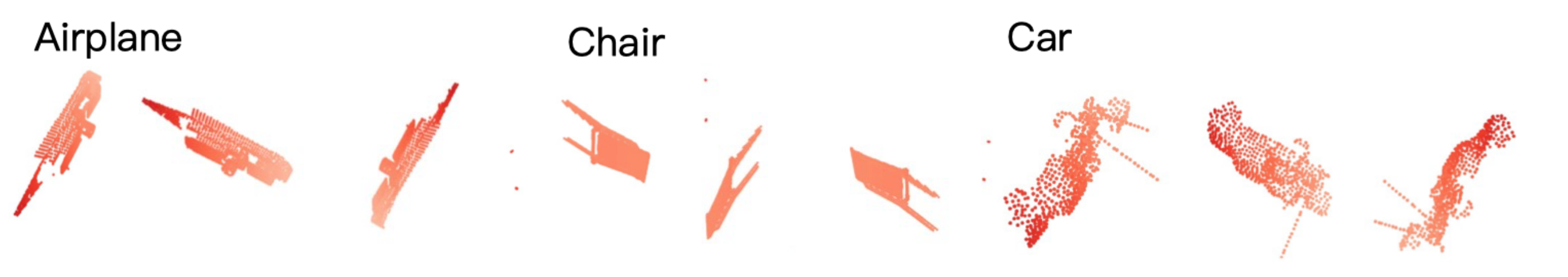}
     \caption{Examples from ShapeNet Occluded which fail to depict the underlying object shapes}
     \label{fig:shapenet_failure}
\end{figure}

\begin{figure}[H]
    \centering
    \includegraphics[width=\linewidth]{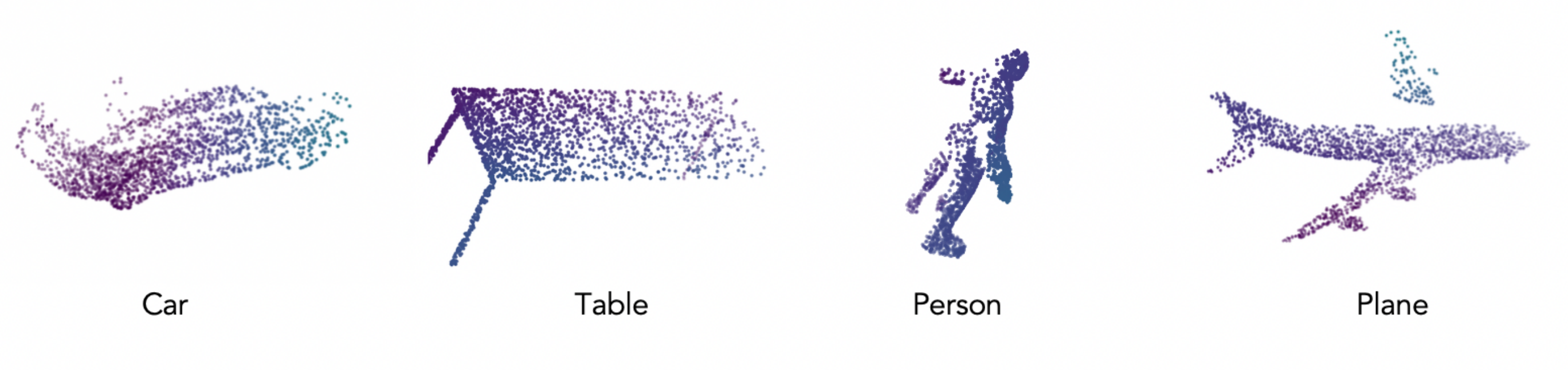}
    \vspace{-2ex}
    \caption{Examples of our generated self-occluded objects from ModelNet.}
    \label{fig:dataoverview}
\end{figure}

\begin{table}[H]
\centering
\caption{Statistics of occluded datasets for \abbrev pre-training}
\begin{tabular}{c|c|c|c|c}
\bottomrule
Name & \# of Class & \# of Object & \# of Views & \# of Points/Object \\ \hline
ShapeNet Occluded (PCN and follow-ups) & 8  & 30974 & 8  & 1045      \\ \hline
ModelNet Occluded (\abbrev)   & 40 & 12304 & 10 & 20085    \\ \toprule
\end{tabular}
\label{tab:occo_stats}
\end{table}
By visualising the objects from the both datasets in Figure~\ref{fig:shapenet_failure} and Figure~\ref{fig:dataoverview}, we show that our generated occluded shapes are more naturalistic and closer to real collected data. We believe this realism will be beneficial for the pre-training. We then test our hypothesis by pre-training models on one of the dataset, and fine tune them on the other. We report these results in Table~\ref{table:occo-dataset}. Clearly we see that the \abbrev models pre-trained on ShapeNet Occluded do not perform as well as the ones pre-trained on ModelNet Occluded in most cases. Therefore we choose our generated ModelNet Occluded rather than ShapeNet Occlude~\cite{yuan2018pcn,Tchapmi_2019_CVPR,WangAL20} used in for the pre-training.

\begin{table}[H]
\centering
\caption{Performance of \abbrev pre-trained models with different pre-trained datasets}
\begin{tabular}{c|c|c|c}\bottomrule
\multicolumn{2}{c|}{\abbrev Settings} & \multicolumn{2}{c}{Classification Accuracy} \\\toprule\bottomrule
Encoder & Pre-Trained Dataset & ModelNet Occ. & ShapeNet Occ. \\\hline
\multirow{2}{*}{PointNet} & ShapeNet Occ. & 81.0 & 94.1 \\\cline{2-4}
 & ModelNet Occ. & \textbf{85.6} & \textbf{95.0} \\\hline
\multirow{2}{*}{PCN} & ShapeNet Occ. & 81.6 & 94.4 \\\cline{2-4}
 & ModelNet Occ. & \textbf{85.1} & \textbf{95.1} \\\hline
 \multirow{2}{*}{DGCNN} & ShapeNet Occ. & 86.7 & 94.5 \\\cline{2-4}
 & ModelNet Occ. & \textbf{89.1} & \textbf{95.1} \\\toprule
\end{tabular}
\label{table:occo-dataset}
\end{table}

\paragraph{Re-Implementation details of "Jigsaw" pre-training methods}
We describe how we reproduce the 'Jigsaw' pre-training methods from \cite{sauder2019self}. Following their description, we first separate the objects/chopped indoor scenes into $3^3=27$ small cubes and assign each point a label indicting which small cube it belongs to. We then shuffle all the small cubes, and train a model to make a prediction for each point. We reformulate this task as a 27-class semantic segmentation, for the details on the data generation and model training, please refer to our released code.

\section{Ablations}
As pointed out by the reviewers, we agree adding more runs will help. To help judge significance, we have ran 10 runs for three settings\footnote{We chose settings that have low FLOPs across tasks and encoders.} 
and computed p-values via t-tests (unpaired, unequal variances) between OcCo and baselines (i.e., Jigsaw or random). We observe that all p-values are below the conventional significance threshold $\alpha \!=\! 0.05$ (the family-wise error rate is also, using Holm-Bonferroni).
\begin{table}[t!]
\setlength{\tabcolsep}{4pt}
\caption{P-values for unpaired (unequal variance) t-tests between OcCo and baselines (across 10 runs).~Setting:~(1) Few-Shot (10-way 10-shot), ScanObjectNN, DGCNN; (2) Classification, ScanNet, PCN; (3) Segmentation, SensatUrban, PointNet.
}
\label{tab:t_test}
\vspace{-2ex}
\begin{center}
\begin{tabular}{ccc}
\toprule
Setting  &  OcCo vs. Rand & OcCo vs. Jigsaw \\ 
\midrule
(1) & $10^{-7}$ & $10^{-7}$ \\ 
(2) & $0.02$ & $0.05$ \\
(3) & $0.006$ & $0.02$ \\
\bottomrule
\end{tabular}
\end{center}
\end{table}

As suggested by the reviewers, we ran ablations varying the number of object views and categories in Tables~\ref{tab:view} and~\ref{tab:cat}. We use Setting (1) from Table~\ref{tab:t_test} as it is the fastest to run ($*$ indicates few-shot result in main paper).

\begin{table}[h]
\setlength{\tabcolsep}{4pt}
\caption{Ablation: number of views (5 runs), $*\!=\!$ main paper result.}
\label{tab:view}
\vspace{-2ex}
\begin{center}
\begin{tabular}{lcccc}
\toprule
\# of Views & 1 & 5 & 10* & 20 \\
\midrule
PointNet & 44.7$\pm$1.8 & 53.6$\pm$1.2 & 54.9$\pm$1.2 & 54.8$\pm$1.0\\
DGCNN    & 42.7$\pm$2.1 & 56.9$\pm$1.4 & 56.8$\pm$1.5 & 57.0$\pm$1.6\\
\bottomrule
\end{tabular}
\end{center}
\end{table}

\begin{table}[h]
\caption{Ablation: number of object categories (5 runs)}
\label{tab:cat}
\vspace{-2ex}
\begin{center}
\begin{tabular}{lccc}
\toprule
\# of Categories\,\,\,\,\,\, & 1 & 10 & 40* \\
\midrule
PointNet & 41.1$\pm$1.2 & 52.2$\pm$1.5 & 54.9$\pm$1.2 \\
DGCNN    & 37.9$\pm$3.8 & 44.8$\pm$2.9 & 56.8$\pm$1.5 \\
\bottomrule
\end{tabular}
\end{center}
\vspace{-6ex}
\end{table}

\section{More results}
\paragraph{3D object classification with Linear SVMs}
We follow the similar procedures from~\cite{achlioptas2017learning,han2019view,sauder2019self,wu2016learning,yang2018foldingnet}, to train a linear Support Vector Machine (SVM) to examine the generalisation of \abbrev encoders that are pre-trained on the occluded objects from ModelNet40. For all six classification datasets, we fit a linear SVM on the output 1024-dimensional embeddings of the train split and evaluate it on the test split. Since~\cite{sauder2019self} have already proven their methods are better than the prior, here we only systematically compare with theirs. We report the results\footnote{In our implementation, we also provide an alternative to use grid search to find the optimal set of parameters for SVM with a Radial Basis Function (RBF) kernel. In this setting, all the \abbrev pre-trained models have outperformed the random and Jigsaw initialised ones by a large margin as well.} in Table~\ref{tab:svm-classification}, we can see that all \abbrev models achieve superior results compared to the randomly-initialised counterparts, demonstrating that \abbrev pre-training helps the generalisation both in-domain and cross-domain. 
\begin{table}[H]
\centering
\caption{linear SVM on the output embeddings from random, Jigsaw and OcCo initialised encoders}
\begin{tabular}{c|c|c|c|c|c|c|c|c|c}
\bottomrule \multirow{2}{*}{Dataset} & 
\multicolumn{3}{c|}{PointNet} & \multicolumn{3}{c|}{PCN} & \multicolumn{3}{c}{DGCNN} \\ \cline{2-10}
 & Rand & Jigsaw & \abbrev& Rand & Jigsaw & \abbrev & Rand & Jigsaw & \abbrev \\ \toprule \bottomrule
ShapeNet10   & 91.3 & 91.1 & \textcolor{blue}{93.9} & 88.5 & 91.8 & \textbf{94.6} & 90.6 & 91.5 & \textcolor{blue}{94.5} \\\hline
ModelNet40   & 70.6 & 87.5 & \textcolor{blue}{88.7} & 60.9 & 73.1 & \textcolor{blue}{88.0} & 66.0 & 84.9 & \textbf{89.2} \\\hline
ShapeNet Oc  & 79.1 & 86.1 & \textcolor{blue}91.1 & 72.0 & 87.9 & \textcolor{blue}90.5 & 78.3 & 87.8 & \textbf{91.6} \\\hline
ModelNet Oc  & 65.2 & 70.3 & \textcolor{blue}{80.2} & 55.3 & 65.6 & \textbf{83.3} & 60.3 & 72.8 & \textcolor{blue}{82.2} \\\hline
ScanNet10    & 64.8 & 64.1 & \textcolor{blue}{67.7} & 62.3 & 66.3 & \textbf{75.5} & 61.2 & 69.4 & \textcolor{blue}{71.2} \\\hline
ScanObjectNN & 45.9 & 55.2 & \textcolor{blue}69.5 & 39.9 & 49.7 & \textcolor{blue}{72.3} & 43.2 & 59.5 & \textbf{78.3} \\\toprule
\end{tabular}
\label{tab:svm-classification}
\end{table}

\paragraph{Few-shot learning}
We use the same setting and train/test split as cTree~\cite{sharma2020self}, and report the mean and standard deviation across on 10 runs. The top half of the table reports results for eight randomly initialised point cloud models, while the bottom-half reports results on two models across three pre-training methods. We bold the best results (and those whose standard deviation overlaps the mean of the best result). It is worth mentioning cTree~\cite{sharma2020self} pre-trained the encoders on both datasets before fine tuning, while we only pre-trained once on ModelNet40. The results show that models pre-trained with OcCo either outperform or have standard deviations that overlap with the best method in 7 out of 8 settings.
\begin{table}[H]
\centering
\setlength{\tabcolsep}{3pt}
\caption{More results on few-shot learning.}
\begin{tabular}{l|cccccccc}
\bottomrule \multicolumn{1}{c|}{\multirow{3}{*}{Baseline}} & 
\multicolumn{4}{c}{ModelNet40} & \multicolumn{4}{c}{Sydney10} \\ \cline{2-9}
& \multicolumn{2}{c}{5-way} & \multicolumn{2}{c}{10-way} & \multicolumn{2}{c}{5-way} & \multicolumn{2}{c}{10-way}\\ \cline{2-9}
& 10-shot & 20-shot & 10-shot & 20-shot & 10-shot & 20-shot & 10-shot & 20-shot \\ \toprule \bottomrule
3D-GAN, Rand & 55.8$\pm$10.7 & 65.8$\pm$9.9 &40.3$\pm$6.5 &48.4$\pm$5.6 & 54.2$\pm$4.6 & 58.8$\pm$5.8 & 36.0$\pm$6.2 & 45.3$\pm$7.9\\
FoldingNet, Rand & 33.4 $\pm$13.1 & 35.8$\pm$18.2 & 18.6$\pm$6.5 & 15.4$\pm$6.8 & 58.9$\pm$5.6 & 71.2$\pm$6.0 & 42.6$\pm$3.4 & 63.5$\pm$3.9\\
Latent-GAN, Rand & 41.6$\pm$16.9 & 46.2$\pm$19.7 & 32.9$\pm$9.2 & 25.5$\pm$9.9 & 64.5$\pm$6.6 & 79.8$\pm$3.4 & 50.5$\pm$3.0 & 62.5$\pm$5.1\\
PointCapsNet, Rand & 42.3$\pm$17.4 &53.0$\pm$18.7 &38.0$\pm$14.3 & 27.2$\pm$14.9 & 59.4$\pm$6.3 & 70.5$\pm$4.8 & 44.1$\pm$2.0 & 60.3$\pm$4.9\\
PointNet++, Rand & 38.5$\pm$16.0 & 42.4$\pm$14.2 & 23.1$\pm$7.0 & 18.8$\pm$5.4 & \textbf{79.9$\pm$6.8} & {85.0$\pm$5.3} & 55.4$\pm$2.2 & 63.4$\pm$2.8\\
PointCNN, Rand & {65.4$\pm$8.9} & {68.6$\pm$7.0} & {46.6$\pm$4.8} & {50.0$\pm$7.2} & 75.8$\pm$7.7 & 83.4$\pm$4.4 & 56.3$\pm$2.4 & {73.1$\pm$4.1}\\
\toprule \bottomrule
PointNet, Rand & 52.0$\pm$12.2 & 57.8$\pm$15.5 & 46.6$\pm$13.5 & 35.2$\pm$15.3 & 74.2$\pm$7.3 & 82.2$\pm$5.1 & 51.4$\pm$1.3 & 58.3$\pm$2.6\\
PointNet, cTree & 63.2$\pm$10.7 & 68.9$\pm$9.4 & 49.2$\pm$6.1 & 50.1$\pm$5.0 &  76.5$\pm$6.3 & 83.7$\pm$4.0 & 55.5$\pm$2.3 & 64.0$\pm$2.4\\
PointNet, OcCo & \textbf{89.7$\pm$6.1} & \textbf{92.4$\pm$4.9} & \textbf{83.9$\pm$5.6} & \textbf{89.7$\pm$4.6} & {77.7$\pm$8.0} & {84.9$\pm$4.9} & {60.9$\pm$3.7} & {65.5$\pm$5.5}\\\hline
DGCNN, Rand & 31.6 $\pm$9.0 & 40.8$\pm$14.6 &19.9$\pm$6.5 & 16.9$\pm$4.8 & 58.3$\pm$6.6 & 76.7$\pm$7.5 & 48.1$\pm$8.2 &76.1$\pm$3.6\\
DGCNN, cTree & 60.0$\pm$8.9 & 65.7$\pm$8.4 & 48.5$\pm$5.6 & 53.0$\pm$4.1 & \textbf{86.2$\pm$4.4} & \textbf{90.9$\pm$2.5} & \textbf{66.2$\pm$2.8} & \textbf{81.5$\pm$2.3}\\
DGCNN, OcCo & \textbf{90.6$\pm$2.8} & \textbf{92.5$\pm$6.0} & \textbf{82.9$\pm$4.1} & \textbf{86.5$\pm$7.1} & \textbf{79.9$\pm$6.7} & \textbf{86.4$\pm$4.7} & 63.3$\pm$2.7 & \textbf{77.6$\pm$3.9}\\
\toprule
\end{tabular}
\end{table}

\paragraph{Detailed results of the part segmentation}
Here in Table~\ref{tab:partseg} we report the detailed scores on each individual shape category from ShapeNetPart, we bold the best scores for each class respectively. We show that for all three encoders, OcCo-initialisation has achieved better results over two thirds of these 15 object classes.
\begin{table}[H]
\centering
\caption{Detailed Results on Part Segmentation Task on ShapeNetPart}
\begin{tabular}{c|c|c|c|c|c|c|c|c|c}
\bottomrule \multirow{2}{*}{Shapes} 
& \multicolumn{3}{c|}{PointNet} 
& \multicolumn{3}{c|}{PCN} 
& \multicolumn{3}{c}{DGCNN} \\ \cline{2-10}
 & Rand* & Jigsaw & \abbrev& Rand & Jigsaw & \abbrev & Rand* & Jigsaw* & \abbrev \\ \toprule \bottomrule
mean (point)& 83.7 & 83.8 & 84.4 & 82.8 & 82.8 & 83.7 & 85.1 & 85.3 & \textbf{85.5} \\\hline
Aero        & 83.4 & 83.0 & 82.9 & 81.5 & 82.1 & 82.4 & 84.2 & 84.1 & \textbf{84.4}\\\hline
Bag         & 78.7 & 79.5 & 77.2 & 72.3 & 74.2 & 79.4 & 83.7 & \textbf{84.0} & 77.5\\\hline
Cap         & 82.5 & 82.4 & 81.7 & 85.5 & 67.8 & \textbf{86.3} & 84.4 & 85.8 & 83.4\\\hline
Car         & 74.9 & 76.2 & 75.6 & 71.8 & 71.3 & 73.9 & 77.1 & 77.0 & \textbf{77.9}\\\hline
Chair       & 89.6 & 90.0 & 90.0 & 88.6 & 88.6 & 90.0 & 90.9 & 90.9 & \textbf{91.0}\\\hline
Earphone    & 73.0 & 69.7 & 74.8 & 69.2 & 69.1 & 68.8 & 78.5 & \textbf{80.0} & 75.2\\\hline
Guitar      & 91.5 & 91.1 & 90.7 & 90.0 & 89.9 & 90.7 & 91.5 & 91.5 & \textbf{91.6}\\\hline
Knife       & 85.9 & 86.3 & 88.0 & 84.0 & 83.8 & 85.9 & 87.3 & 87.0 & \textbf{88.2}\\\hline
Lamp        & 80.8 & 80.7 & 81.3 & 78.5 & 78.8 & 80.4 & 82.9 & 83.2 & \textbf{83.5}\\\hline
Laptop      & 95.3 & 95.3 & 95.4 & 95.3 & 95.1 & 95.6 & 96.0 & 95.8 & \textbf{96.1}\\\hline
Motor       & 65.2 & 63.7 & 65.7 & 64.1 & 64.7 & 64.2 & 67.8 & \textbf{71.6} & 65.5\\\hline
Mug         & 93.0 & 92.3 & 91.6 & 90.3 & 90.8 & 92.6 & 93.3 & 94.0 & \textbf{94.4}\\\hline
Pistol      & 81.2 & 80.8 & 81.0 & 81.0 & 81.5 & 81.5 & \textbf{82.6} & \textbf{82.6} & 79.6\\\hline
Rocket      & 57.9 & 56.9 & 58.2 & 51.8 & 51.4 & 53.8 & 59.7 & \textbf{60.0} & 58.0\\\hline
Skateboard  & 72.8 & 75.9 & 74.2 & 72.5 & 71.0 & 73.2 & 75.5 & \textbf{77.9} & 76.2 \\\hline
Table       & 80.6 & 80.8 & 81.8 & 81.4 & 81.2 & 81.2 & 82.0 & 81.8 & \textbf{82.8} \\\hline
\end{tabular}
\label{tab:partseg}
\end{table}

\newpage



\newpage
\section{Algorithmic Description of OcCo}
\begin{lrbox}{\codebox}
\begin{lstlisting}
# P: an initial point cloud
# K: camera intrinsic matrix
# V: number of total view points
# loss: a loss function between point clouds
# c: encoder-decoder completion model
# p: downstream prediction model

while i < V:
    # sample a random view-point
    R_t = [random.rotation(), random.translation()]
    
    # map point cloud to camera reference frame
    P_cam = dot(K, dot(R_t, P))
    
    # create occluded point cloud
    P_cam_oc = occlude(P_cam, alg='z-buffering')
    
    # point cloud back to world frame
    K_inv   = [inv(K), zeros(3,1); zeros(1,3), 1]
    R_t_inv = transpose([R_t; zeros(3,1), 1])
    P_oc = dot(R_t_inv, dot(K_inv, P_cam_oc))

    # complete point cloud
    P_c = c.decoder(c.encoder(P_oc))
    
    # compute loss, update via gradient descent
    l = loss(P_c, P)
    l.backward()
    update(c.params)
    i += 1

# downstream tasks, use pre-trained encoders
p.initialise(c.encoder.params)
p.train()
\end{lstlisting}
\end{lrbox}

\begin{algorithm}[H]
  \caption{Occlusion Completion (OcCo)}\label{alg:occo}
  \usebox{\codebox}
\end{algorithm}

\newpage
\section{Visualisation from Completion Pre-Training}
\begin{figure}[H]
    \centering
    \includegraphics[width=\textwidth]{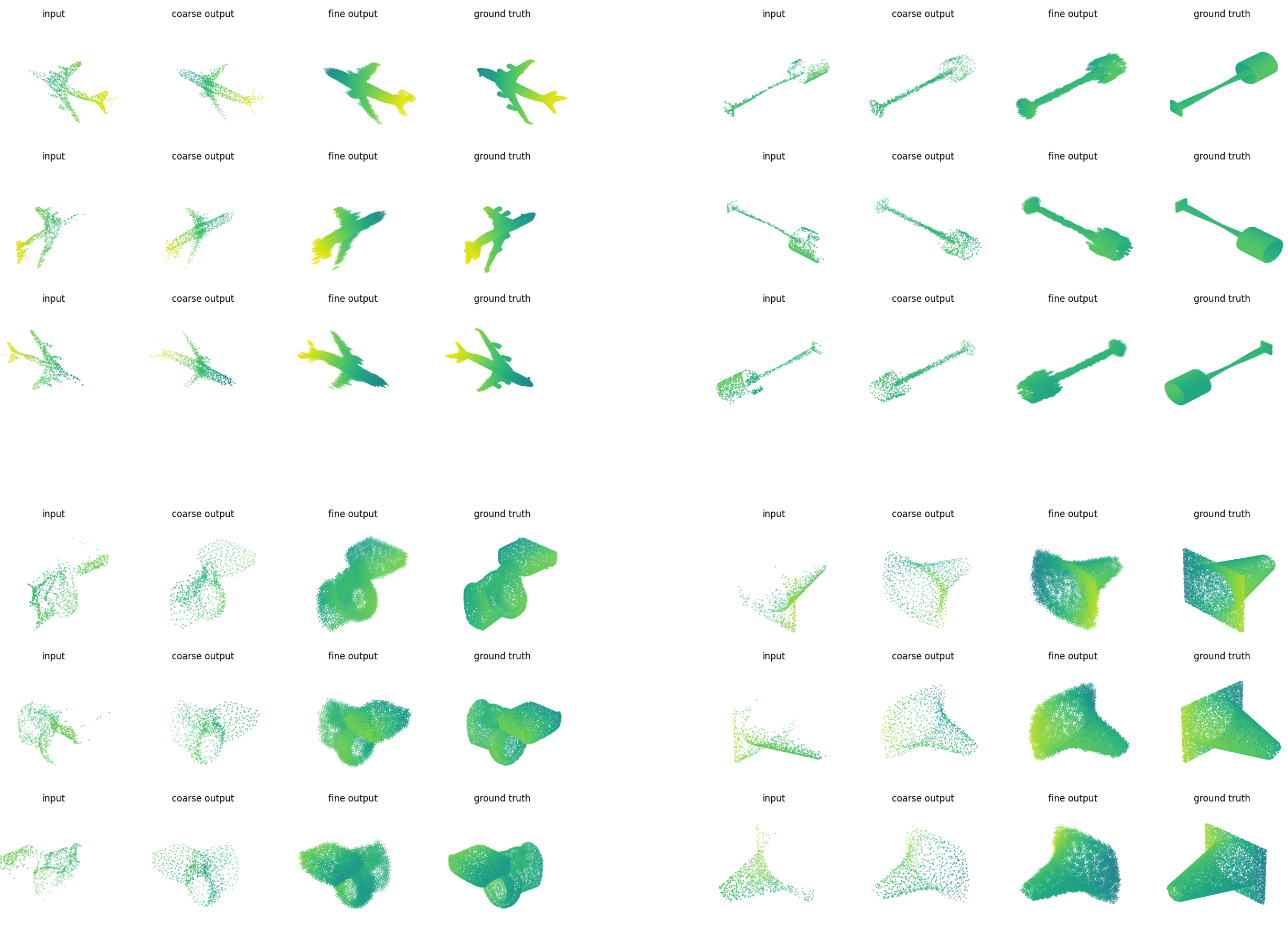}
    \caption{OcCo pre-training with PCN encoder on occluded ModelNet40.}
    \label{fig:qualitative_occo_pcn}
\end{figure}

\begin{figure}[H]
    \centering
    \includegraphics[width=\textwidth]{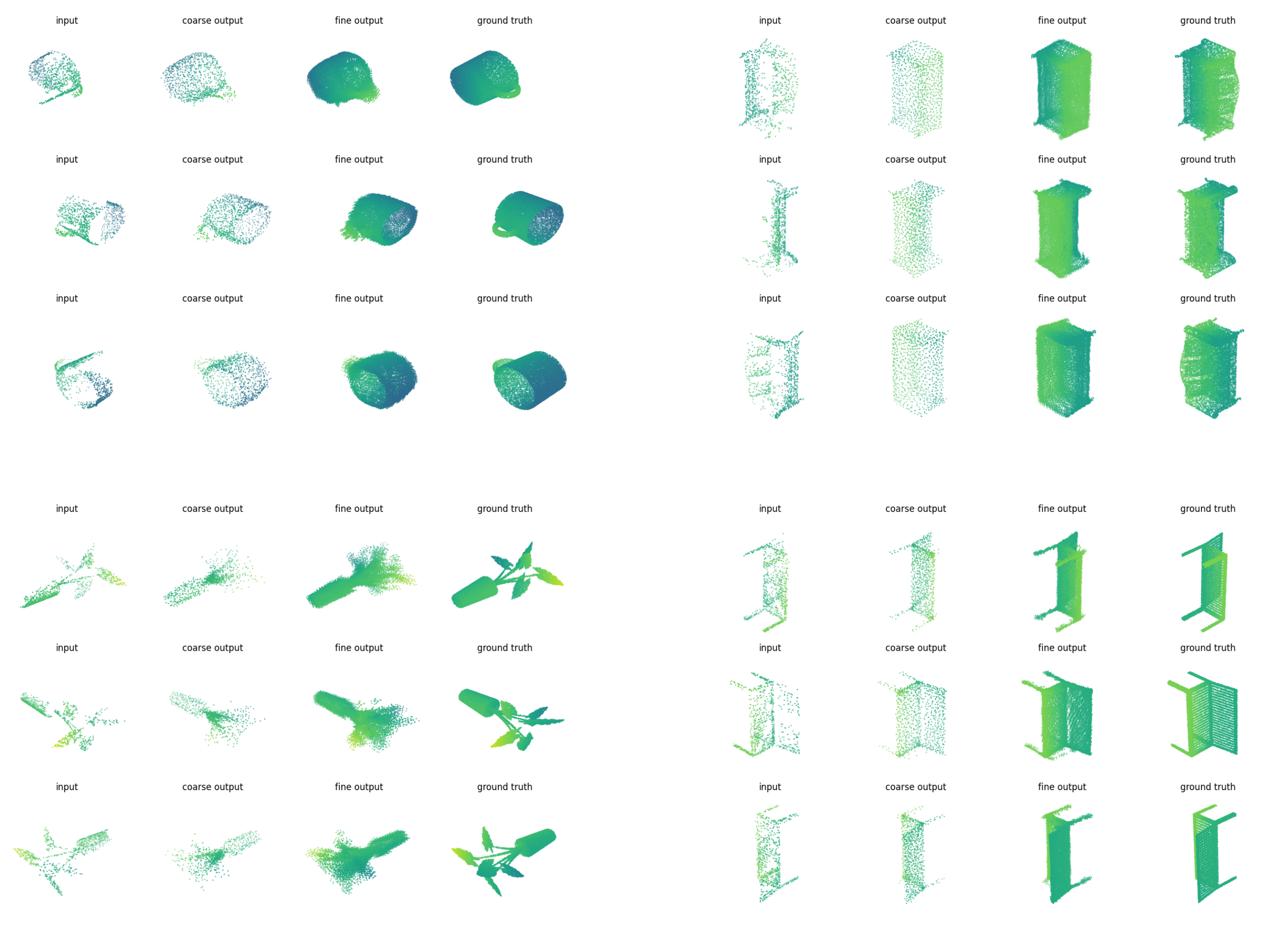}
    \caption{OcCo pre-training with PointNet encoder on occluded ModelNet40.}
    \label{fig:qualitative_occo_pointnet}
\end{figure}

\begin{figure}[H]
    \centering
    \includegraphics[width=\textwidth]{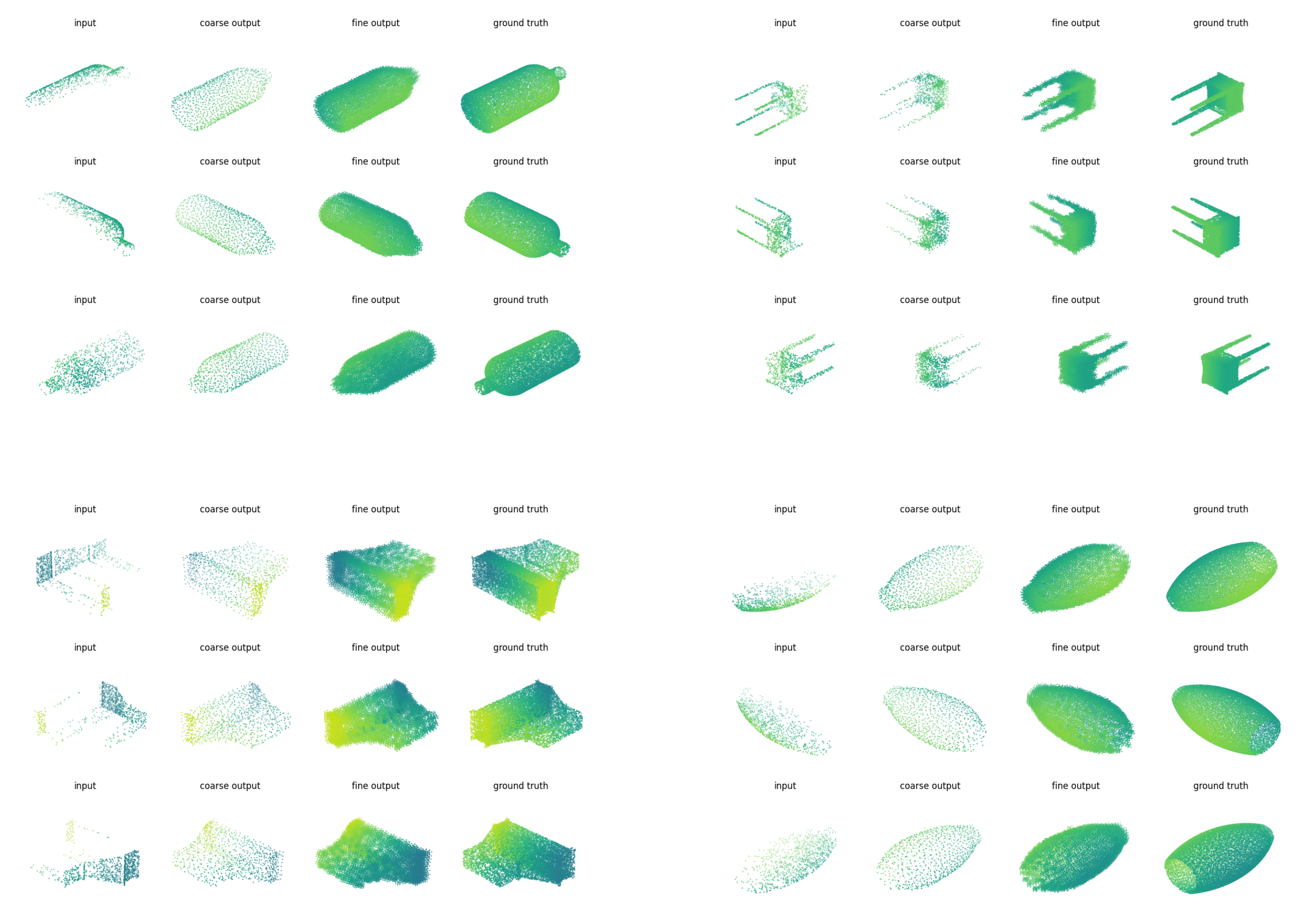}
    \caption{OcCo pre-training with DGCNN encoder on occluded ModelNet40.}
    \label{fig:qualitative_occo_dgcnn}
\end{figure}

\begin{figure}[H]
    \centering
     \includegraphics[width=\textwidth]{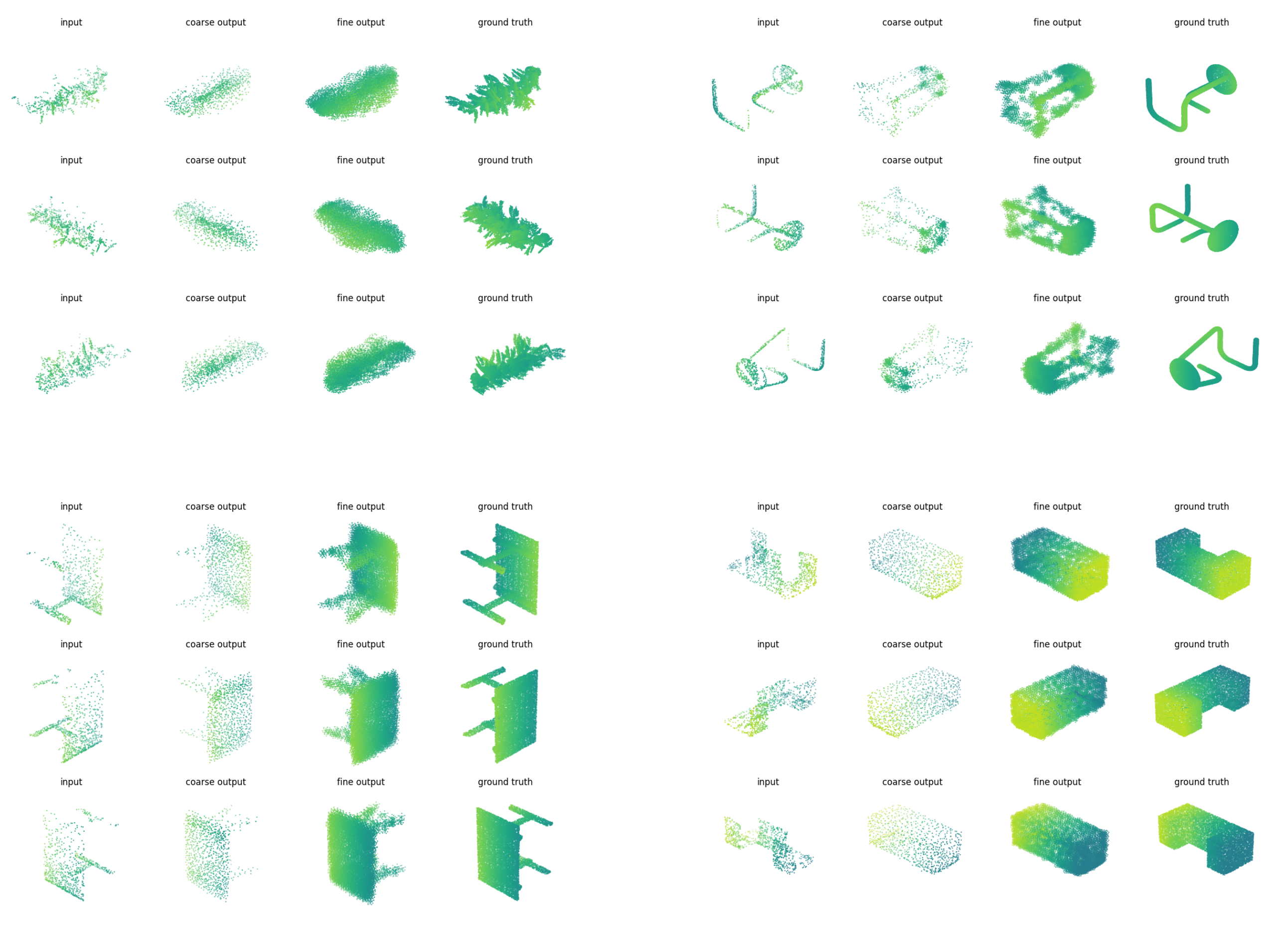}
     \caption{Failure completed examples during OcCo pre-training.}
     \label{fig:failuare_occo}
\end{figure}

In this section, we show some qualitative results of \abbrev pre-training by visualising the input, coarse output, fine output and ground truth at different training epochs and encoders. In Figure.~\ref{fig:qualitative_occo_pcn}, Figure.~\ref{fig:qualitative_occo_pointnet} and Figure.~\ref{fig:qualitative_occo_dgcnn}, we notice that the trained completion models are able to complete even difficult occluded shapes such as plants and planes.
In Figure.~\ref{fig:failuare_occo} we plot some failure examples of completed shapes, possibly due to their complicated fine structures, while it is worth mentioning that the completed model can still completed these objects under the same category.
\noindent
\newpage
\noindent
\newpage
\noindent


\end{document}